\documentclass{article}
\usepackage[utf8]{inputenc}
\usepackage[english]{babel}

\usepackage{natbib}
\setcitestyle{authoryear,open={(},close={)}} 

\providecommand{\keywords}[1]
{
  \small	
  \textbf{\textit{Keywords---}} #1
}

\title{A model aggregation approach for high-dimensional large-scale optimization}
\author{Haowei Wang$^{1}$, Ercong Zhang$^{1}$, Szu Hui Ng$^{1}$, Giulia Pedrielli$^{2}$  \\
        \small $^{1}$National University
of Singapore, $^{2}$Arizona State University\\
}
\date{}

\usepackage{natbib}
\usepackage{graphicx}
\usepackage{amsmath}
\usepackage{amsfonts}
\usepackage{amssymb}
\usepackage{mathtools}
\usepackage{commath}
\usepackage{bm}
\usepackage[ruled,linesnumbered]{algorithm2e}
\SetKwComment{Comment}{$\triangleright$\ }{}
\usepackage{listings}
\usepackage{graphicx} 
\usepackage{float} 
\usepackage{subfigure} 
\usepackage{setspace}
\usepackage{verbatim}
\usepackage{color}
\usepackage{diagbox}
\usepackage{appendix}
\usepackage{geometry}
\usepackage[colorlinks,
    linkcolor=red,
    anchorcolor=blue,
    citecolor=green
]{hyperref}

\usepackage{amsthm}
\newtheorem{example}{Example}
\newtheorem{theorem}{Theorem}
\newtheorem{lemma}{Lemma}
\newtheorem{assumption}{Assumption}
\theoremstyle{remark}
\newtheorem{remark}{Remark}
\theoremstyle{definition}
\newtheorem{definition}{Definition}
\DeclareMathOperator{\E}{\mathbb E}

\def\P{\mathbb P}
\def\M{\mathcal M}

\begin{document}

\maketitle

\begin{abstract}
Bayesian optimization (BO) has been widely used in machine learning and simulation optimization.  With the increase in computational resources and storage capacities in these fields, high-dimensional and large-scale problems are becoming increasingly common.  In this study, we propose a model aggregation method in the Bayesian optimization (MamBO) algorithm for efficiently solving high-dimensional large-scale optimization problems.  MamBO uses a combination of subsampling and subspace embeddings to collectively address high dimensionality and large-scale issues; in addition, a model aggregation method is employed to address the surrogate model uncertainty issue that arises when embedding is applied.  This surrogate model uncertainty issue is largely ignored in the embedding literature and practice, and it is exacerbated when the problem is high-dimensional and data are limited.  Our proposed model aggregation method reduces these lower-dimensional surrogate model risks and improves the robustness of the BO algorithm.  We derive an asymptotic bound for the proposed aggregated surrogate model and prove the convergence of MamBO.  Benchmark numerical experiments indicate that our algorithm achieves superior or comparable performance to other commonly used high-dimensional BO algorithms.  Moreover, we apply MamBO to a cascade classifier of a machine learning algorithm for face detection, and the results reveal that MamBO finds settings that achieve higher classification accuracy than the benchmark settings and is computationally faster than other high-dimensional BO algorithms.
\end{abstract}


\section{Introduction}
Global optimization problems are of immense interest in various fields, including machine learning, computer science, and engineering. Many problems can be formulated as global optimization problems, such as hyperparameter tuning problems for machine learning algorithms \citep{snoek2012practical}, the performance optimization of the controller for a physical robot \citep{guzman2020heteroscedastic}, and price optimization problems in revenue management \citep{phillips2021pricing}. Among these, the hyperparameter tuning problem is of considerable interest in machine learning. Recent machine learning models have become increasingly complex with a large set of hyperparameters, and the values of the parameters may have a significant impact on the algorithm's performance. In hyperparameter tuning, we aim to determine a set of hyperparameter values that can achieve the best performance. However, hyperparameter tuning problems still face several challenges because the parameter space is usually high-dimensional, and each evaluation may be extremely computationally 
expensive for complex models. This motivates us to design an efficient global optimization 
algorithm that could potentially be applied to such a problem.

Formally, let us consider the following optimization problem
\begin{equation}
\min_{x\in\mathcal{X}}\E[y(x)], \label{def}
\end{equation}
where $x$ is the input, $\mathcal{X} \subset \mathbb{R}^d$ is the search space, which is usually assumed to be compact, and $y(x)$ is the observed output at $x$. Typically, we assume $y(x)=f(x)+\xi(x)$, where $f(x)$ is the true objective function and $\xi(x)$ represents the stochastic noise, which is normally distributed with mean 0. In this study, we consider a general heteroscedastic noise, that is, the variance of the noise can vary across different inputs.

In problems such as hyperparameter tuning, the objective function $f(x)$ can be viewed as a black-box function, implying that there is no closed-form expression of $f$, and the cost for each evaluation of $f$ may be expensive. Owing to the high cost of these evaluations, it is generally difficult or impossible to estimate the derivative information for $f$. Therefore, the use of traditional gradient-based optimization methods is limited. Several optimization methodologies can be applied to solve \eqref{def} under this black-box, derivative-free setting, including evolutionary algorithms \citep{yu2010introduction}, random search methods \citep{zhigljavsky2012theory}, and surrogate-based simulation optimization (see \cite{hong2021surrogate} for an overview). Each method exhibits unique advantages and shortcomings under different settings. Among these methods, Bayesian optimization (BO), which is a surrogate-based method, has recently been widely used owing to its mathematical convenience and flexibility \citep{frazier2018bayesian}. BO uses the Gaussian process (GP) model as a surrogate with a probabilistic framework that helps query the search space efficiently. Furthermore, BO does not require strict assumptions on the output $y$ and is easy to apply.

In this study, we propose an efficient BO algorithm to solve problem \eqref{def} in a high-dimensional and large-scale setting.

\subsection{Motivation}\label{motivation}
From hyperparameter tuning problems for machine learning or even deep learning models to complex simulation optimization problems for complex industrial systems, the need for an efficient high-dimensional and large-scale optimization algorithm has become ubiquitous. Although there have been many successful applications of BO algorithms, the optimization problem \eqref{def} can still be challenging when the input dimension $d$ (i.e., the dimension of $x$) and number of observations $n$ are large.

As systems become increasingly complex, many real applications require the solution of a high-dimensional optimization problem. For example, \cite{wang2018batched} described a rover trajectory-planning problem in which 30 location points need to be optimized with respect to a reward for the rover trajectory. To optimize such high-dimensional systems, it is difficult to directly apply BO. The number of evaluations required to find the global optimum increases exponentially in $d$, thereby resulting in poor scaling of the BO to high-dimensional problems with limited budgets.

Large-scale systems have also been applied in various fields, for example, optimizing the simulation model for scheduling semiconductor factories \citep{hildebrandt2014large}, which employs a complex simulation model that contains more than 3000 jobs per run. However, for large-scale problems (large $n$), a computational effort of $\mathcal{O}(n^3)$ in time is required to obtain the posterior predictive distribution of the GP model in BO (because we need to compute the inverse of the covariance matrix). This cubic time complexity makes the BO not scale well when $n$ increases to the thousands \citep{williams2006gaussian}.

Moreover, problems with large $d$ and large $n$ may occur simultaneously as the computational capacity and storage capacity increase, for example, in the parameter tuning task for large-scale machine learning algorithms. This type of task is essential as well as challenging, as many of those algorithms today contain a large number of hyperparameters to tune, where the different values of hyperparameters can significantly impact the performance of such algorithms. For example, the Viola \& Jones Cascade (VJ) classifier \citep{viola2001rapid} is a machine learning algorithm that detects whether a given image contains a face, and the aim is to optimize its classification accuracy. As there are more than 20 parameters to tune in the VJ classifier for optimization, and several hundreds of iterations are needed to make the optimization procedure converge, this can be considered a high-dimensional large-scale optimization problem (a more detailed description of this problem is mentioned in Section \ref{sec:vj}). In such cases, it is of interest to address the large $d$ and large $n$ problems simultaneously to optimize the models more efficiently.

Owing to these challenges, high-dimensional large-scale optimization algorithms can be designed to solve \eqref{def}. In this study, we also consider this high-dimensional large-scale scenario and propose an efficient BO algorithm to solve the optimization problem given by \eqref{def}. In the next section, we review the existing BO algorithms designed to solve either the high dimensional or large scale scenarios, and the few designed for both. 

\subsection{Related Work}\label{sec:relatedwork}
In this subsection, we introduce related literature that addresses the dimensionality (large $d$) and scalability (large $n$) issues mentioned in Section \ref{motivation}.

There are two main streams of work for solving problem \eqref{def} with a large $d$. The first assumes a low-dimensional intrinsic structure of the objective function $f$. \cite{wang2016bayesian} proposed the REMBO algorithm, which assumes that $f$ only varies along a lower dimensional subspace, and GP-EI is performed in a stochastic subspace generated via random embedding. \cite{djolonga2013high} implemented a variant of REMBO with a detailed analysis of the regret with the upper confidence bound acquisition function. 
Many other extensions of REMBO were subsequently proposed later; for example, \citep{binois2015warped,nayebi2019framework,letham2020re,cartis2022dimensionality}.
The second stream assumes an additive structure on $f$: \cite{kandasamy2015high} proposed the Add-GP-UCB algorithm, in which $f$ is considered as a summation of functions, each of which only depends on a disjoint subset of dimensions. \cite{li2016high} extended the idea of Add-GP-UCB to non-disjoint cases. In addition to these two main streams, several other methods have been proposed to address high dimensionality. In \cite{oh2018bock},  cylindrical kernels in the GP are adopted to transform the geometry of the search space to avoid extensive searching at the boundary. The method of \cite{li2017high} adopts a dropout technique, so that only a subset of variables is optimized in each iteration. \cite{eriksson2021high} proposed the sparse axis-aligned subspace Bayesian optimization (SAASBO) algorithm, which 
places a hierarchical sparse prior over the kernel parameters of the GP model to ensure that most of the unimportant dimensions are ``turned off'' during the optimization procedure.

Separately, several studies have attempted to address the large $n$ issue. One common approach is to select candidate design points in batches such that $f$ can be simulated in parallel \citep{marmin2015differentiating,gonzalez2016batch,wang2020parallel}. Another technique involves the use a sparse GP when constructing a GP model. The sparse GP uses an additional set of inducing points to approximate the stochastic GP model. \cite{nickson2014automated} and \cite{mcintire2016sparse} adopted sparse GP in the BO framework. 
Recently, \cite{meng2022combined} proposed a combined global and local search for optimization (CGLO) algorithm, where the optimization procedure is built upon the additive global and local GP (AGLGP) model. This AGLGP model originates from the sparse GP and combines the modelling of a global trend and several local trends in different local regions. 
Moreover, several studies have replaced the GP model with other surrogates, such as a neural network \citep{snoek2015scalable}, to improve the scalability to a large number of observations.

Furthermore, some studies have aimed to address both the large $d$ and large $n$ problems simultaneously. \cite{wang2018batched} uses an ensemble of additive GP models to build the ensemble Bayesian optimization (EBO) algorithm. The additive structure makes the algorithm search more efficient in a high-dimensional space. It also adopts a blocked approximation of the covariance matrix with a parallel query that scales the EBO to deal with large-scale problems with tens of thousands of observations. \cite{eriksson2019scalable} proposed the trust region Bayesian optimization (TuRBO) algorithm, which uses a set of trust regions that are centered around the current best to provide an improved local modeling. This local modeling strategy with random restart makes the TurBO scale effective for large $n$ and large $d$ problems by focusing only in a small region at each iteration.

\subsection{Illustration and Contributions}
Although several BO algorithms have been designed for high-dimensional large-scale problems, the computational time for non-embedding-based methods is typically much longer than that for embedding-based methods. This issue becomes more severe when the budget is limited. Therefore, in this study, we develop an embedding-based algorithm to solve the above problem.

The majority of existing studies that apply dimension reduction through embedding assume that the number of active dimensions is known. However, this is unrealistic in practice, because the active dimensions are typically unknown. Furthermore, in many of the  proposed algorithms, it is further assumed that, based on embedding onto an unknown active dimension from a limited set of data, the resulting embedded model is considered as the “true” model that describes the high-dimensional process. When the dataset is small and limited, especially at the beginning of the iterative algorithms, the embedding or projection method is highly sensitive and dependent on the dataset, resulting in high uncertainty in the “best” embedded model to describe the high dimensional process. Applying a single embedded model from a small dataset ignores this uncertainty and can potentially misinform the high-dimensional process, and decisions that are made during the process can be suboptimal. In Example \ref{motivatingexample}, we illustrate the impact of ignoring the  embedding uncertainty in an optimization problem for a simplified 100 dimensional example.

\begin{example}\label{motivatingexample}
Suppose that the objective function $f$ is the Hartman-6 function embedded in a 100 dimensional space with the search space $\mathcal{X}=[0,1]^{100}$, and the noisy response of $f$ obtained is denoted as $y$. We collect five independent datasets (i.e. five different initial designs) from this process, each with 20 input locations and observations taken at those locations. Then, for each dataset, we apply PCA embedding onto six dimensions  and proceed to optimize the function using a straightforward BO (with the EI acquisition function) in this lower-dimensional space.
\end{example}

\begin{table}[h]
\begin{tabular}{|c|c|c|c|c|c|c|c|}
\hline
                   & dim 1  & dim 2  & dim 3  & dim 4  & dim 5  & dim 6  & simple regret \\ \hline
trial 1            & 0.1334 & 0.1265 & 0.2380 & 0.0009 & 0.3624 & 0.8090 & 2.2432        \\ \hline
trial 2            & 0.2205 & 0.3672 & 0.3668 & 0.0971 & 0.3310 & 0.6597 & 1.3160        \\ \hline
trial 3            & 0.2134 & 0.3049 & 0.2999 & 0.3861 & 0.2422 & 0.4781 & 1.4737        \\ \hline
trial 4            & 0.5000 & 0.1119 & 0.4844 & 0.4953 & 0.2538 & 0.8224 & 2.0502        \\ \hline
trial 5            & 0.2452 & 0.2508 & 0.0933 & 0.1727 & 0.1923 & 0.6838 & 1.5086        \\ \hline
true minimal point ($x^*$) & 0.2017 & 0.1500 & 0.4769 & 0.2753 & 0.3117 & 0.6573 & -             \\ \hline
\end{tabular}
\caption{Performance for Hartman-6 function with five different initial designs}
\label{example}
\end{table}

\begin{figure}[htbp] 
\centering 
\includegraphics[width=0.6\textwidth]{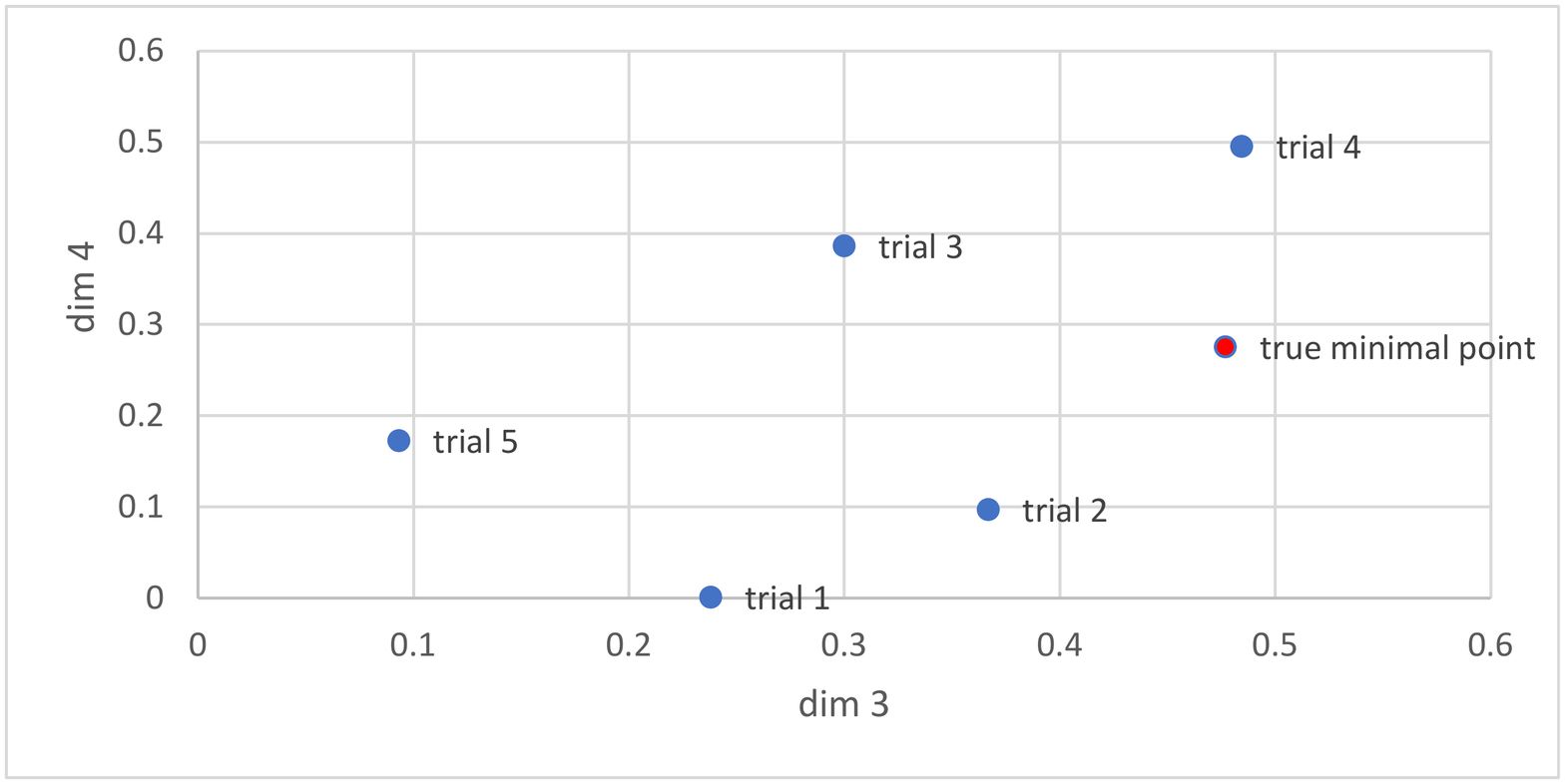} 
\caption{Scatter plot of $\hat{x}^*$ and $x^*$ in dim 3 and dim 4} 
\label{example_dim34} 
\end{figure}

The optimization results for the five different models from the five datasets are shown in Table \ref{example}. In Table \ref{example}, each row shows the returned minimal point ($\hat{x}^*$) in the six original active dimensions for a single trial, with the last row showing the true minimal point ($x^*$) in the original six active dimensions, and the last column provides the simple regret for 5 different trials. Here, the simple regret is defined as the gap between the returned minimum and true minimum ($|f(\hat{x}^*)-f(x^*)|$), and the returned minimal points ($\hat{x}^*$) for the five models are obtained after 80 iterations. From Table \ref{example}, we see that each model from each of the 5 datasets returns a different final optimum point $\hat{x}^*$ and responses $f(\hat{x}^*)$. Figure \ref{example_dim34} plots the different
$\hat{x}^*$ and $x^*$ for active dimensions 3 and 4, and it illustrates how varied the estimated optimal points can be depending on the data sampled, and how they can be suboptimal when the sample size is limited. For all 5 datasets, there is a gap between $\hat{x}^*$ and $x^*$.
 
The results from this simple example clearly demonstrate that although all five datasets are drawn from the same underlying process and the same embedding method is applied, the resulting lower-dimensional model may differ, and more importantly, the decisions drawn from these models may be suboptimal.

To reduce the impact of embedding uncertainty on our decisions, as illustrated in Example \ref{motivatingexample}, in this study, we propose to mitigate this risk by developing a model aggregation and optimization approach. In our approach, we first apply subsampling to the dataset to train a more robust set of submodels and then use a Bayesian model aggregation approach to combine the models.

The main contributions of this study can be summarized as follows:
\begin{enumerate}
    \item We propose a Bayesian aggregated model
    that utilizes a subsampling technique and subspace embedding to reduce computational effort when dealing with large-scale and high-dimensional problems. This model generalizes the earlier work proposed in \cite{xuereb2020stochastic} from an averaged predictor to an aggregated Gaussian process. More importantly, we adopt a model aggregation framework to reduce the embedding uncertainty, which has been largely ignored in previous studies.
    We also provide an asymptotic bound for the predictor of the Bayesian aggregated model to control the information loss due to subspace embedding.
    
    \item With this Bayesian aggregated model, we further propose a model aggregation method in Bayesian optimization (MamBO) algorithm to solve problem \eqref{def} with large $d$ and large $n$. We also provide theoretical proof of the convergence of this algorithm.
    
    \item  We finally provide a numerical comparison against state-of-the-art benchmark algorithms. We also apply our approach to a well-known real problem in computer vision for face detection to illustrate the practical use of our approach. 
\end{enumerate}

The advantages of such a modeling approach are twofold. First, subsampling and embedding greatly facilitate model development for large $d$ and $n$. Second, model aggregation mitigates embedding uncertainty and enhances overall model robustness and predictive performance. With this aggregated model, a BO algorithm is proposed to address large $d$ and large $n$.

The remainder of this paper is organized as follows: In Section \ref{background}, we provide an overview of the stochastic GP model and BO. In Section \ref{algorithm}, we introduce the Bayesian aggregated model and propose the MamBO algorithm in detail. In Section \ref{proof}, we present the convergence analysis of the MamBO algorithm. In Section \ref{experiment}, we conduct a numerical comparison of MamBO with the most commonly used benchmark algorithms. Section \ref{conclusion} concludes the paper.

\section{Background}\label{background}
\subsection{Stochastic GP Model}
In this study, we assume that the observed output of a black-box function can be modeled with the stochastic GP model in the following form:
\begin{equation}
Y(x)=F(x)+\xi(x),
\end{equation}
where $Y(x)$ represents the GP for the output $y(x)$, $F(x)$ represents the GP for the objective function $f(x)$, and $\xi(x)$ is the GP for heteroscedastic noise with mean 0. We treat $f(x)$ as a realization of a GP $F(x)$ with mean $l(\cdot)^T\beta$ and covariance function $\Sigma_F(\cdot,\cdot)$, where $l(\cdot)$ is a vector of known basis functions. If we further assign a prior distribution for $\beta$ to $\mathcal{N}(b,\Omega)$, then the stochastic model for $f(x)$ becomes
\begin{equation}
F(x)\sim \mathcal{GP}(l(x)^Tb,l(x)^T\Omega l(x')+\Sigma_F(x,x'))
\end{equation}
This model helps capture the prior information on the objective function \citep{xie2014bayesian}.

For an optimization problem with stochastic noise, replicates are typically taken at each input $x_i$ for $i=1,\cdots,n$. Let $\bar{Y}(x_i)$ be the sample mean and $s^2(x_i)$ be the sample variance of all replicates at $x_i$. For any input point $x$, the posterior distribution of $F(\cdot)$ is given by: 
\begin{equation}
    F(x)\mid \bar{Y}(x_1),\cdots,\bar{Y}(x_n) \sim \mathcal{GP}(m(x),C(x,x')),
\end{equation}
where $m(x)$ is the conditional mean function 
\begin{equation}
    m(x)=l(x)\hat{\beta}+\Sigma_F^T(x,\cdot)\left[\Sigma_F+\Sigma_{\xi}\right]^{-1}(\bar{Y}-L^T\hat{\beta}),
\end{equation}
and $C(x,x')$ is the conditional covariance function
\begin{equation}
\begin{aligned}
    C(x,x')=&\Sigma_F(x,x')-\Sigma_F(x,\cdot)^T\left[\Sigma_F+\Sigma_{\xi}\right]^{-1}\Sigma_F(x',\cdot)\\
    &+u(x)^T\left[\Omega^{-1}+L(\Sigma_F+\Sigma_{\xi})^{-1}L^T\right]^{-1}u(x'),
\end{aligned}
\end{equation}
where $\Sigma_F(x,\cdot)$ is the covariance vector between $x$ and other input points, $\Sigma_F$ is the $n\times n$ covariance matrix of $(F(x_1),\cdots,F(x_n))$, $\Sigma_\xi$ is the  $n\times n$ covariance matrix of $(\xi(x_1),\cdots,\xi(x_n))$, $\bar{Y}=\left[\bar{Y}(x_1),\cdots,\bar{Y}(x_n)\right]^T$, $L=\left[l(x_1),\cdots,l(x_n)\right]$,  $\hat{\beta}=\left[\Omega^{-1}+L(\Sigma_F+\Sigma_{\xi})^{-1}L^T)\right]^{-1}$, and $u(x)=l(x)-L(\Sigma_F+\Sigma_{\xi})^{-1}\Sigma_F(x,\cdot)$. 

In practice, we must also specify the form of $\Sigma_F(x,x')$ in advance. For example, we can use $\Sigma_F(x,x')=\sigma_F^2 R_F(x,x')$, where $R_F(x,x')=\sum_{i=1}^d\exp\{-\theta_i(x_i-x_i')^2\}$ is the squared exponential correlation function (also known as the kernel function). For the kernel parameters $\theta_i$ and process variance $\sigma_F^2$, we can also place a conjugate prior on them, which makes the resulting posterior predictive distribution a non-central student-$t$ distribution \citep{ng2012bayesian}. In such cases, we can still use a stochastic GP model as a reasonable approximation of the unknown response surface when the degree of freedom of the posterior predictive model tends to infinity.

\subsection{Bayesian Optimization (BO)}\label{sec:bo}
In this subsection, we briefly introduce the BO framework (see \cite{frazier2018bayesian} for a comprehensive review). 
We summarize the general BO algorithm for noisy observations in Algorithm \ref{BO}. 

There are two key components of BO: the surrogate model and the acquisition function. The GP model is typically used as the surrogate, which provides an approximation of the unknown objective function, and the acquisition function is the sampling criterion to guide the location of the next observation. The concept of BO is as follows: Based on the current observations, we first construct a GP surrogate model that reflects our current belief in the objective function. The acquisition function is then applied to help determine the next point $x\in\mathcal{X}$ where we obtain a new observation. Once we obtain a new observation, we can update the GP model iteratively. As the number of iterations increases, the GP model approximation of the objective function improves. In the final step, the input point with the lowest sample mean is reported as the final solution.

    \begin{algorithm}
        \KwIn{Acquisition function $\alpha$}
        Sample the initial points using some initial design of experimental criteria, such as Latin hypercube design, and obtain the initial observations $\mathcal{D}_0=\left\{\left(x_1,\bar{y}(x_1)\right),\cdots,\left(x_{n_0},\bar{y}(x_{n_0})\right)\right\}$
        
        Fit an initial GP model with $\mathcal{D}_0$ and use cross-validation to check whether this initial model is valid
        
        \For{$i=0,1,\cdots,I$}{
            Set $x_{n_0+i+1}=\arg\max\alpha(x|\mathcal{D}_i)$\;
            Data augmentation: $\mathcal{D}_{i+1}=\mathcal{D}_i \cup\{(x_{n_0+i+1},\bar{y}(x_{n_0+i+1}))\}$\;
            Update the GP with $\mathcal{D}_{i+1}$\;

        } 
        \KwRet{The sampled point with the lowest sample mean}
        \caption{Bayesian Optimization for noisy observations}
        \label{BO}
    \end{algorithm}
    
The general algorithm described in Algorithm \ref{BO} is applicable to both the noisy and noiseless cases. For the noiseless case, $\bar{y}(x_i)$ is a single observation taken at $x_i$, and the final value returned by the algorithm is the point with the lowest observation.

The purpose of the acquisition function in the algorithm is to determine the next point to evaluate, and it is typically designed to balance exploration and exploitation \citep{frazier2018bayesian}. The popular choices for the acquisition function include  expected improvement (EI) \citep{jones1998efficient}, upper/lower confidence bounds \citep{Srinivas2010GaussianPO}, Thompson sampling (\citet{thompson1933likelihood} \& \citet{russo2018tutorial} for a review), entropy search \citep{hennig2012entropy}, and knowledge gradient \citep{frazier2009knowledge}.

\section{Model Aggregation Method in Bayesian Optimization (MamBO)}\label{algorithm}
As we consider the case when both $n$ and $d$ are large, it will be challenging to adopt the regular BO (Algorithm \ref{BO}) directly, as the modeling and update of GP will be computational infeasible.  Many previous works solve the high dimensional problem by adopting an embedding matrix to reduce the dimensionality. This type of embedding-based method works by utilizing a subspace embedding to the data, which results in a lower dimensional model, and applying BO only on the single embedded model. Here, lines 1 and 6 are modified in Algorithm 1 to include an embedding procedure on the data collected before the GP model is (re)-fitted with the lower dimensional data.  However, as illustrated in Example \ref{motivatingexample}, when the initial data set is small or data is noisy, the embedded model can be inaccurate, and hence when using it to drive the search and optimization, it can result in suboptimal results (as shown in Table \ref{example}). Further, although the embedding approach does overcome the challenge with large $d$, it however still does not address the computational issues when $n$ gets large.

In this section, we propose the model aggregation method in Bayesian optimization (MamBO) algorithm to solve problem (\ref{def}) when $n$ and $d$ are large, that also accounts for the model uncertainty when using embedding to drive the BO.

\subsection{Overview of MamBO}
The key idea of MamBO is to modify the stochastic GP model in Algorithm \ref{BO}, such that the resulting optimization algorithm can efficiently deal with the dimensionality and scalability issues, and also the embedding uncertainty. Following this idea, we propose a Bayesian aggregated model (which will be discussed in detail in Section \ref{model}) in this study.
This Bayesian aggregated model also belongs to the class of GP model, which leverages on subspace embedding and subsampling to address the dimensionality and scalability issues of large $n$ and $d$, and a model aggregating approach to account for the embedding uncertainty. More specifically, to address the large $n$ issue, we first divide the data into different subsets, and train a separate GP model for each subset (with the use of embedding to address large $d$).  As each of these models is plausible based on the data sampled, a Bayesian approach is then taken to combine the individual GP models together into an aggregated model that can better account for the model uncertainty due to the limited data and uncertainty in embedding. Once this Bayesian aggregated model is built in each iteration, a searching criterion can be applied to continue the optimization procedure (line 4 in Algorithm \ref{BO}).

Overall, MamBO adopts the general iterative BO approach in Algorithm \ref{BO}, but adapts for large $n$ and $d$, and also the embedding uncertainty, by developing the Bayesian aggregated model (replacing the standard GP model in lines 2 and 6) to drive the BO search. The details of the MamBO algorithm with its components will be developed in the following subsections.

\subsection{Bayesian Aggregated Model}\label{model}
In this section, we extend the idea of \cite{xuereb2020stochastic} and propose a novel GP surrogate model when $d$ and $n$ are large. In \cite{xuereb2020stochastic}, a stochastic GP model averaging (SGPMA) predictor was proposed to solve high-dimensional large-scale prediction problem of some noisy black-box function $f(x)$. The SGPMA predictor adopts the subspace embedding and subsampling technique to tackle the dimensionality and scalability issues (which will be discussed in detail later), and then uses the Bayesian model averaging approach (an overview is provided in \cite{fragoso2018bayesian}) to combine the predictor of the subsampled lower dimensional GP models. However, in the context of global optimization, a GP surrogate model is needed instead of a single predictor. As the Bayesian model averaging approach focuses on the posterior distribution of the observations, instead of the process itself, here we propose to build a novel GP model in a model aggregation framework as the surrogate in BO to solve problem (\ref{def}). To address the dimensionality and scalability issues, we still adopt the subspace embedding and subsampling approach as in \cite{xuereb2020stochastic}. Furthermore, to mitigate the risk of embedding uncertainty with limited data, instead of using a single embedded model as the surrogate in BO, here we further propose a Bayesian aggregated model in lieu of a single “best” stochastic GP model. More specifically, random embedding is applied to solve the large $d$ problem, subsampling is applied to solve the large $n$ problem, and finally, model aggregation is applied to reduce the embedding uncertainty when applying these techniques on a finite dataset. To outline, the initial sampled points with observations are divided into subsets, and on each subset, we train a GP model defined on a random projected subspace. The resulting model returned by the Bayesian aggregated model is a weighted average of each submodel. The weights are selected to be the posterior model weights, and they can be updated throughout the optimization procedure as the observations increase. We show later that the predictor of the Bayesian aggregated model is asymptotically bounded in Theorem \ref{thm:bound}, which ensures that information loss due to random embedding can be controlled. 
Algorithm \ref{BM} presents the procedure for the Bayesian aggregated model. Parameters $m$, $n_i$, and $d_i$ are selected by random sampling.

    \begin{algorithm}
        \KwIn{Design matrix $X=\left[x_1^T,\cdots,x_n^T\right]^T \in \mathbb{R}^{n\times d}$ with observed sample mean $\bar{Y} \in \mathbb{R}^n$; the number of subsets to be generated $m$ }
        Randomly divide $x_1,\cdots,x_n$ into $m$ subgroups and let $X_i\in \mathbb{R}^{n_i\times d}$ be the resulting new design matrix for subgroup $i$ with the corresponding observation $Y_i\in\mathbb{R}^{n_i}$
        
        \For{$i=1,\cdots,m$}{
            $X_i \leftarrow \Pi_i(X_i)$ where $\Pi_i$ is a subspace embedding for group $i$ to dimension $d_i$, $d_i$ is the reduced dimension\; \Comment{For example, $\Pi_i$ can be a random embedding \citep{xuereb2020stochastic} or a PCA embedding}
            Build a stochastic GP model $\mathcal{M}_i\sim \mathcal{GP}(m_i,C_i)$ for $f(x)$ using $X_i,Y_i$\;
            Compute the Bayes weight $w_i(X_i)$ for each $\mathcal{M}_i$\;
        } 
        \KwRet{The aggregated model $F_n=\sum_{i=1}^mw_i(X_i)\M_i$ together with its conditional mean $\mu_n=\sum_{i=1}^mw_i(X_i)m_i$ and conditional covariance function $k_n=\sum_{i=1}^mw_i(X_i)^2C_i$} 
        \caption{Bayesian aggregated model}
        \label{BM}
    \end{algorithm}

The idea of model aggregation is to aggregate submodels that rely only on a subset of data, with each submodel being faster to compute. More precisely, we first construct $m$ GP models $\M_1,\cdots,\M_m$, with each $\M_i$ built from a subset $X_i\in \mathbb{R}^{n_i \times d}$ of the design matrix $X$. Because $n_i<<n$, the computational cost for constructing each $\M_i$ is greatly reduced compared with the direct construction of a GP with all data. We then combine the submodels $\M_1,\cdots,\M_m$ to obtain an aggregated model:
\begin{equation}
F_n=\sum_{i=1}^mw_i(X_i)\M_i,
\end{equation}
\label{BAM}
where the weights $w_i$ satisfy $w_i \geq 0$ and $\sum_{i=1}^m w_i =1$.

The aggregated model $F_n$ is still a GP because it is a finite sum of GPs, and the GPs are closed under linear operations \citep{williams2006gaussian}. Suppose each submodel $\M_i\sim \mathcal{GP}(m_i,C_i)$; then, according to \cite{tanaka2019spatially}, we have $F_n\sim \mathcal{GP}(\mu_n,k_n)$, where $\mu_n(x)=\sum_{i=1}^mw_im_i(x)$, and $k_n(x,x')=\sum_{i=1}^m w_i^2C_i(x,x')$.

\begin{remark}
The conditional mean function (also the predictor) $\mu_n$ of the Bayesian aggregated model is the same as that proposed by \cite{xuereb2020stochastic}. \cite{xuereb2020stochastic} aims to obtain a robust estimator for the predictive mean. In this work, as the model drives the BO search, we focus on obtaining a more computationally efficient GP model $F_n$ of the objective function $f(x)$ and proceed the optimization procedure with this GP model.
\end{remark}

An important component of the aggregated model is the weight $w_i(X_i)$. We consider the Bayes weight in our setting, that is, the weight is chosen to be the posterior model probability $\P(\M_i\mid X_i,Y_i)$, and it is proportional to the product of the model prior and the likelihood function. This Bayes weight considers both our prior knowledge of the model and our current belief and can be continually updated when more observations become available. We describe our weight estimation procedure in Algorithm \ref{weight}.

    \begin{algorithm}
        \KwIn{GP model $\mathcal{M}_i$, design matrix $X_i$, observation $Y_i$, a parameter $\eta$}
        
        \For{$i=1,\cdots,m$}{
            Set the model prior for each $\mathcal{M}_i$, $$\P(\mathcal{M}_i)= \frac{\left(\frac{n_i}{n}\right)^2\left(\frac{d_i}{d}\right)^\eta}{\sum_{j=1}^m\left(\frac{n_j}{n}\right)^2\left(\frac{d_j}{d}\right)^\eta};$$
            
            Compute the Bayes weight $$w_i(X_i)=\P(\mathcal{M}_i \mid X_i,Y_i)=\frac{\P(Y_i \mid \mathcal{M}_i)\P(\mathcal{M}_i)}{\sum_{j=1}^m\P(Y_j \mid \mathcal{M}_j)\P(\mathcal{M}_j)};$$
        } 
        \caption{Bayes Weight Estimation}
        \label{weight}
    \end{algorithm}

For the choice of model prior $\P(\M_i)$, a commonly used choice in literature is a vague prior (i.e. $\P(\M_i)\propto 1$), which assumes that we have no prior knowledge on which model is better, and the model posterior will be proportional to the marginal likelihood. However, model characteristics, such as model size, will affect the model prediction power; hence, a uniform assumption in our case is not reasonable. Here we adopt the same model prior as in \cite{xuereb2020stochastic}, $\P(\M_i)\propto\left(\frac{n_i}{n}\right)^2\left(\frac{d_i}{d}\right)^\eta$. This model prior includes the information of both the number of observations $n_i$ used and the reduced dimension $d_i$. The terms $\frac{n_i}{n}$ and $\frac{d_i}{d}$ measure the fraction of the $i$-th sample size and dimension compared with the whole dataset, respectively, and the parameter $\eta$ adjusts the relative importance of the dimension compared with the sample size in each submodel. Because $\eta$ is an unknown hyperparameter, we propose setting it using cross-validation. (Specifically, $\eta$ is selected from a discretized set of size $k$, and we choose the best $\eta$ with the lowest simple regret using $k$-fold cross-validation.) Moreover, to efficiently compute $P(Y_i\mid \M_i)$, we adopt the well-known BIC approximation \citep{konishi2008information}, in which the error caused by this approximation is $O_p(1)$ \citep{wasserman2000bayesian}.

\begin{remark}
Other common choices for $w_i(X_i)$ include uniform weights, likelihood values, and some information criterion-based weights such as BIC weights \citep{burnham2004multimodel}.
Uniform weights crudely treat all the submodels equally and ignore the fitted information. 
Compared to the likelihood weights and BIC weights, the Bayes weights additionally consider an important and intuitive piece of information about the submodels: the more in-sample data and features used for training a submodel, the more weights should be assigned to the submodel. The parameter $\eta$ helps to adjust the weights for different submodels with different model characteristics. Moreover, the Bayes weights can be treated as a generalization of BIC weights. BIC weight is also a posterior model probability when the prior probability of each model is the same.
\end{remark}

\begin{remark}
An important advantage of adopting the Bayesian aggregated model is that the computational complexity is significantly reduced from $\mathcal{O}(n^3)$ to $\mathcal{O}(n_1^3+\cdots+n_m^3)$ with $\sup\{n_1,...,n_m\} \ll n$, compared with the stochastic GP model. 
\end{remark}

At the end of this section, we prove an asymptotic bound for the predictor (also the conditional mean function) $\mu_n$ of the Bayesian aggregated model. This bound ensures that the information loss due to subspace embedding will not be extensive, as the number of iterations tends to infinity. To begin our analysis, we formally define $\epsilon'$-subspace embedding.
\begin{definition}[$\epsilon'$-subspace embedding]
Given a matrix $V$ with orthonormal columns, an embedding $\Pi$ is called an $\epsilon'$-subspace embedding for $V$ if, for $\forall x$,
\[(1-\epsilon')\norm{Vx}_2^2\leq\norm{\Pi Vx}_2^2\leq (1+\epsilon')\norm{Vx}_2^2\]
Or equivalently, $\norm{V^T\Pi^T \Pi V-I}_2\leq \epsilon'$. \citep{paul2014random}
\end{definition}

An example of this $\epsilon'$-subspace embedding is a Gaussian random matrix, where each entry is a rescaled standard normal random variable. 
With this definition, we can conclude the asymptotic property of the predictor in Theorem \ref{thm:bound}. The proof is provided in the appendix.
\begin{theorem}\label{thm:bound}
Suppose the kernel parameters in the Bayesian aggregated model are known, and $\Pi_i$ is an $\epsilon'$-subset embedding for all $i$; then, as $n \rightarrow \infty$, \[\P\left(\left|\mu_n-f(x)\right|\geq \epsilon'\norm{x}_2B\right)\rightarrow 0,\]
where $B$ is a fixed constant.
\end{theorem}

\subsection{Acquisition Function}\label{sec:acquisition}
In this section, we summarize how this Bayesian aggregated model can be applied to commonly used acquisition functions.

Suppose our initial set of input points with corresponding observations is  
 $\mathcal{D}_0=\left\{\left(x_1,\bar{y}(x_1)\right),\ldots,\left(x_{n_0},\bar{y}(x_{n_0})\right)\right\}$, where $\bar{y}(x_i)$ is the mean observed output at $x_i\in\mathbb{R}^d$ for $i=1,\ldots,n_0$. Here, we assume that both $n$ and $d$ are sufficiently large to emphasize the high dimensionality and scalability of the problem. 

 Based on the Bayesian aggregated model, we can use $F_n:=\sum_{i=1}^mw_i\mathcal{M}_i$ as the GP model for the objective function $f(x)$. To proceed with BO using $F_n$, we need to detail the acquisition function we use. 

There are various types of acquisition functions. Among them, EI has been widely applied since \cite{jones1998efficient}. 
The two main advantages of EI are that it has a closed-form expression; hence, it is easier to compute and optimize. Second, EI can automatically balance the trade-offs between exploration (which tends to select design points with a high posterior mean) and exploitation (which tends to select design points with a high posterior variance).
Owing to computational and sample efficiency, we chose EI as an example in this paper. 
EI is defined as the expected value between the current best and posterior GP models, and the detailed form of EI is given by Equation \eqref{ei}.
\begin{equation}
\begin{aligned}
\mathrm{EI}_{T}\left(x\right)&=\E\left[(T-F_{n}(x))^+\mid\mathcal{D}_n\right]\\
&=\Delta \Phi\left(\frac{\Delta}{\sqrt{k_{n}\left(x,x\right)}}\right)+\sqrt{k_{n}\left(x,x\right)} \phi\left(\frac{\Delta}{\sqrt{k_{n}\left(x,x\right)}}\right)
,
\label{ei}
\end{aligned}
\end{equation}
where $T=\min\{\bar{y}(x_1),\cdots,\bar{y}(x_{n_0+n})\}$ is the approximated best current value for $f(x), \Delta=T-\mu_{n}(x)$. 
For more details of the EI acquisition function, please refer to \cite{jones1998efficient}. 

In addition to EI, other acquisition functions can be used, such as upper confidence bound (UCB) \citep{Srinivas2010GaussianPO} and Thompson sampling \citep{thompson1933likelihood}. 
The UCB acquisition function is defined based on the statistical bound of the GP model, and it includes information on both the posterior mean and posterior variance of the GP. In Thompson sampling, we simply draw a random sample $\Tilde{f}(x)$ of the current GP posterior as the acquisition function, which is computationally efficient.

\subsection{MamBO}
With the proposed Bayesian aggregated model, we detail our proposed MamBO algorithm to solve problem (\ref{def}) with large $n$ and $d$. In Table \ref{tab:notation}, we define the key notations used in MamBO, and an outline of the MamBO algorithm is summarized in Algorithm \ref{mambo}.

\begin{table}[h]
\begin{center}
\begin{tabular}{|l|l|}
\hline
     & \multicolumn{1}{c|}{Definition}                         \\ \hline
$N$    & total budget                                            \\ \hline
$n_0$    & size of the initial space filling design                \\ \hline
$A$    & remaining number of replications                \\ \hline
$i$    & current iteration                                       \\ \hline
$B_i$    & available number of replications for iteration $i$                \\ \hline
$r_{\text{min}}$ & minimum number of replications for any new point \\ \hline
$\mathcal{D}_i^X$ & Set of all sampled points at iteration $i$ \\ \hline
\end{tabular}
\end{center}
\caption{Key notations for MamBO}
\label{tab:notation}
\end{table}

    \begin{algorithm}
        \KwIn{$\mathcal{D}_0=\left\{\left(x_1,\bar{y}(x_1)\right),\cdots,\left(x_{n_0},\bar{y}(x_{n_0})\right)\right\}$, some general acquisition function $\alpha(x)$}
        Fit an initial Bayesian aggregated model (\ref{BAM}) with $\mathcal{D}_0$ and use cross-validation to check whether this initial model is valid.\\
        Set $i\leftarrow 0,A\leftarrow N-n_0\cdot r_{\text{min}}$\;
        \While{$A>0$}{
        \underline{Search stage}: select $x_{n_0+i+1}=\arg\max_{x\in\mathcal{X}-\mathcal{D}_{i}^X}\alpha(x)$\;
        \underline{Observe} $(x_{n_0+i+1},\bar{y}(x_{n_0+i+1}))$ with $r_{\min}$ replications and augment $\mathcal{D}_{i+1}=\mathcal{D}_{i}\cup \{(x_{n_0+i+1},\bar{y}(x_{n_0+i+1}))\} $\;
        \underline{Allocation stage}: decide $B_i$ and allocate additional replicates in sampled points\;
        \underline{Build} $F_{i+1}\sim\mathcal{GP}(\mu_{i+1},k_{i+1})$ using the Bayesian aggregated model with $\mathcal{D}_{i+1}$\;
        $A\leftarrow A-r_{\text{min}}-B_i,i\leftarrow i+1$\;
            
        }
        \KwRet{The sampled point with the lowest sample mean}
        \caption{MamBO}
        \label{mambo}
    \end{algorithm}

At each iteration $i$, the next point to be observed is determined by the acquisition function $\alpha(x)$ and a noisy response is observed at that point. The training dataset is then augmented with this new observation, and the Bayesian aggregated model is updated. As new outputs are observed, different submodels are trained from different subsets of the augmented data. This can mitigate the risks of continuously using some of the earlier models (where fewer data are observed), which can have a poorer fit and are less accurate. The weight of each submodel is also updated in each round, which helps reduce the embedding uncertainty in the model.
 
 To address heteroscedastic noise, we add an additional allocation stage to Algorithm \ref{mambo}. This method was first proposed in \cite{quan2013simulation} and further modified in \cite{pedrielli2020extended}. It aims to reduce the uncertainty due to the variability at the sampled points. To guarantee the convergence of MamBO, we make the following assumption of the noise and the allocation rule. Essentially, this assumption requires that we allocate sufficient budget to each design point in the first $i$ iterations. 
 \begin{assumption}
The noise variance function $\sigma^2_{\xi}(x)$ is bounded, that is, $\max_{x\in \mathcal{X}}\sigma^2_{\xi}(x) <\infty$. Additionally, there exists a sequence $\{s_i\}$ such that $s_{i+1}\geq{s_i}$ , $s_i\to \infty$ as $i \to \infty$, and $\sum_{i=1}^\infty i \exp(-as_i)<\infty$ for any positive $a$. The allocation rule ensures that $M_i(x)\geq s_{N_i}$ for the selected points $x$, where $M_i(x)$ is the total number of replications for $x$ in the first $i$ iterations and $N_i$ is the number of points selected in the first $i$ iterations.
\end{assumption}

As seen from lines to 5-6, each iteration budget is distributed between exploration and exploitation and managed by the acquisition function; the search stage and replication evaluation are managed by the allocation stage criteria. In the search stage (line 4), we sample the new design point with $r_{\text{min}}$ replications by optimizing the acquisition function. In the allocation stage (line 6), we further allocate the available budget $B_i$ among the sampled design points. For the choice of $B_i$, we adopted the choice in \cite{pedrielli2020extended} and set $B_i=\sum_{i=1}^{N_i}\max\{0,s_{N_i}-M_i(x)\}$. For the allocation rule, some commonly used rules such as the optimal computing budget allocation (OCBA) technique \citep{chen2000simulation} and equal allocation satisfying Assumption 1 can be applied. In this work, we distribute an additional number of replicates using the OCBA technique \citep{chen2000simulation}. The OCBA rule is given by
 \begin{gather}
     \frac{N_i^{\text{OCBA}}}{N_j^{\text{OCBA}}}=\left(\frac{s(x_i)/d_{b,i}}{s(x_j)/d_{b,j}}\right)^2,i\neq j \\
     N_b^{\text{OCBA}}=s(x_b)\sqrt{\sum_{i\neq b}\frac{(N_i^\text{OCBA})^2}{s^2(x_i)}}
 \end{gather}
 where $x_b$ is the sampled point with the lowest $\bar{y}(x_b)$, $s^2(x_i)$ is the sample variance for $x_i$, $d_{b,i}$ is the distance between $\bar{y}(x_i)$ and $\bar{y}(x_b)$, and $N_i^{\text{OCBA}}$ is the number of new replicates allocated to $x_i$. 
 By adding the allocation stage to the algorithm, we can better manage the number of replications required for each sampled point when faced with heteroscedastic simulation noise.

\section{Convergence Analysis of MamBO}\label{proof}
In this section, we present the convergence results of our proposed algorithm. Here, we apply the EI acquisition function. First, we list the general assumptions.

\begin{assumption}
The objective function $f$ is bounded.
\end{assumption}
\begin{assumption}
The kernel parameters $\theta_i$ and the process variance $\sigma_F^2$ in the Bayesian aggregated model are known.
\end{assumption}
\begin{assumption}
The noise variances $\sigma_{\xi}^2(x)$ are known.
\end{assumption}

Assumptions 2-3 are general assumptions that are also applied in the convergence proof of BO with EI for noiseless outputs \citep{jones1998efficient}. Assumptions 1 and 4 were used in \cite{pedrielli2020extended} for the convergence proof of BO with EI and the allocation step for the stochastic response. The convergence result for MamBO is subsequently stated in Theorem \ref{thm:conv}.

\begin{theorem}
Under Assumptions 1-4, the optimal value returned by MamBO at iteration $t$, $\bar{y}^*_t$, converges to the true global optimum $f(x^*)$ as $t\to \infty$, where $x^*=\arg\min_{x \in \mathcal{X}}{f(x)}$.
\label{thm:conv}
\end{theorem}

To prove Theorem \ref{thm:conv}, we separate the derivation into two steps. The first step is summarized in Lemma 1, where we aim to show that the points visited by MamBO are dense. In the second step, we prove the convergence of MamBO based on Lemma 1.

\begin{lemma}
Under Assumptions 1-4, the sequence of points $\{x_n\}$ visited by MamBO becomes dense in $\mathcal{X}$ as the number of observations $n \rightarrow \infty$.
\end{lemma}

All the proof details can be found in the appendix. 

\begin{remark}
We also note here that although the predictor $\mu_n$ is not a consistent estimator of $f(x)$ in general, the convergence of MamBO is still guaranteed. This is because we return the point with the lowest sample mean in each iteration, and the sample means are normally distributed with mean $f(x)$ and variance depending on the number of replicates at $x$. When the budget increases to infinity, according to Lemma 1, the point sequence visited by MamBO becomes dense in the original $d$-dimensional space. Eventually, all points will be visited. Because we allocate an additional budget to sampled points, the variance of sample means tends to 0 as $n \rightarrow \infty$. As the algorithm proceeds, it follows that all points in the high-dimensional space will be accurately observed. Hence, MamBO can return the optimal function value when the number of iterations approaches infinity.
\end{remark}

\section{Numerical Experiments}\label{experiment}
In this section, we demonstrate the empirical performance of the MamBO algorithm. We compare this with five commonly used algorithms for high-dimensional and large-scale problems, and test them across different test problems. We first test them across a suite of standard test functions typically used in global optimization. Next we test MamBO on a more practical price optimization problem in revenue management. Finally, we also include a real face recognition (image recognition) problem to demonstrate another practical application of MamBO. Both $d$ and $n$ in these problems range from the twenties to hundreds, covering problems on both high dimensional and large scales.

\subsection{Benchmark Algorithms}
We compared our MamBO algorithm with the following common baseline algorithms: REMBO, HesBO, TuRBO, ALEBO, and SAASBO. REMBO \citep{wang2016bayesian} is a high-dimensional BO algorithm that utilizes GP-EI in a randomly embedded subspace. HesBO \citep{nayebi2019framework} is a recently proposed variant of REMBO. Unlike the original REMBO embedding matrix (a random matrix with entries being standard normal distributed), HesBO adopts a sparse count-sketch projection and hence requires less computational effort. HesBO outperformed REMBO in several numerical examples \citep{nayebi2019framework}. ALEBO \citep{letham2020re} is another recent variant of REMBO. Several refinements have been adopted in ALEBO, including the Mahalanobis kernel function and an inequality constraint when optimizing the acquisition function. These refinements are better suited to the structure of the embedded subspace. In addition to these three embedding based algorithms, we also include two non-embedding based algorithms in the benchmark. TuRBO \citep{eriksson2019scalable} is a BO algorithm that is designed for large-scale high-dimensional problems. Instead of building a global GP surrogate model for the entire search space, TuRBO maintains several trust regions centered around the current best and builds local probabilistic models. This helps avoid exploring highly uncertain regions, and by limiting the volume of the search region, TurBO can avoid the curse of dimensionality. SAASBO \citep{eriksson2021high} is a recently proposed algorithm. The main idea to SAASBO is to adopt a variable selection approach. It places a structured sparse prior for the kernel parameters in the GP model, which then ``turns on'' only a subset of important dimensions during the optimization process, facilitating the computations and avoiding the overfitting of the model.

Other algorithms based on additive structures, such as Add-GP-UCB \citep{kandasamy2015high} and EBO \citep{wang2018batched}, lack scalability to large datasets, resulting in extremely long computational times when the input dimension $d$ is large. The computation generally becomes extremely slow when $d$ gets into the low hundreds.\footnote{We used an HPC cluster (CPU E5-2690 v3 @ 2.60 GHz with 24 cores) to run Add-GP-UCB for the Branin function with $d=100$, which terminated after 24 h with only 88 iterations.} Hence, we excluded this from our numerical experiments.

\subsection{Synthetic Data}
\begin{figure}[!htb]
\centering
\subfigure
[Simple regret for 100 dim Branin] 
{\includegraphics[width=6cm]{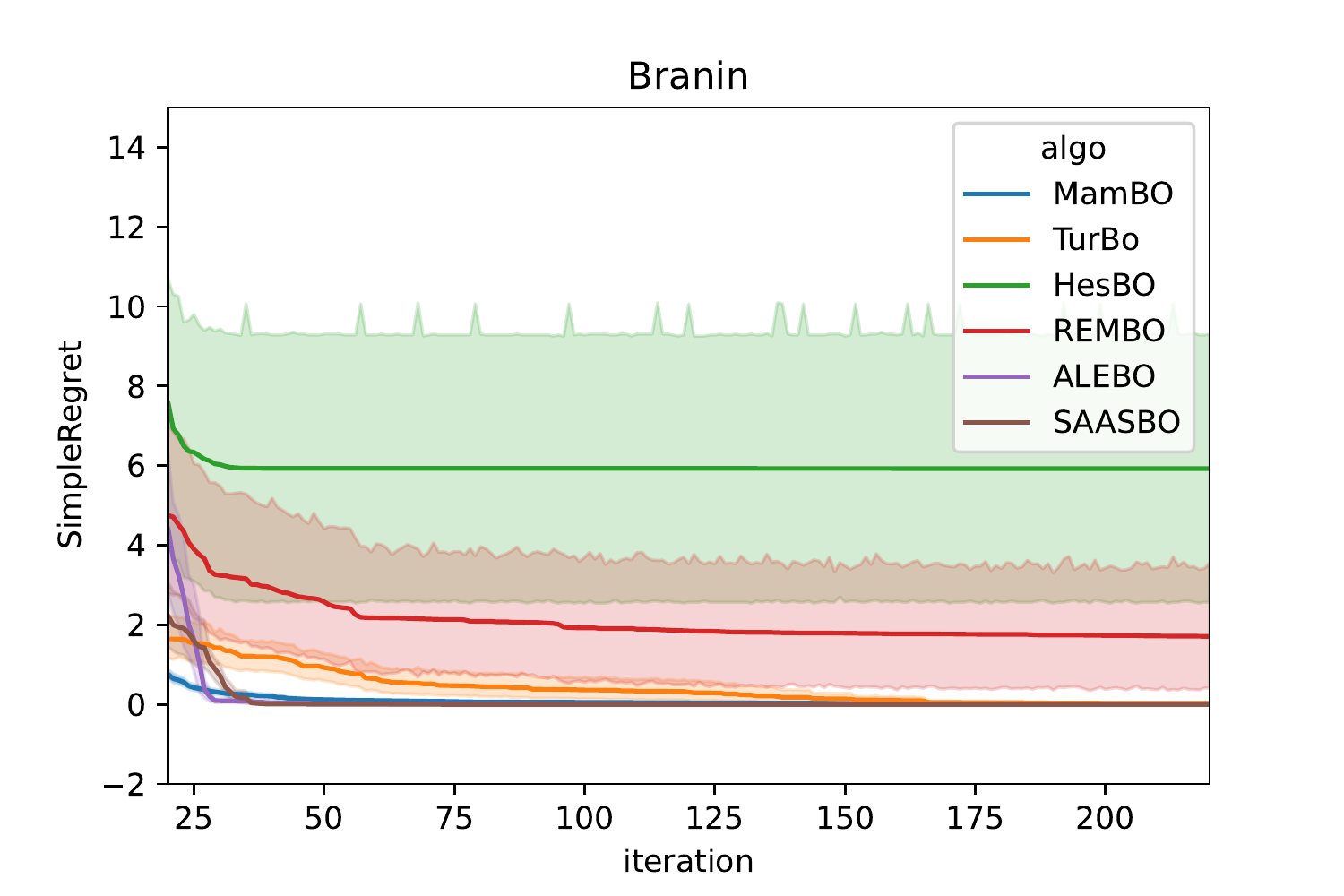}} 
\subfigure[Boxplot of simple regret for 100 dim Branin]{\includegraphics[width=8cm]{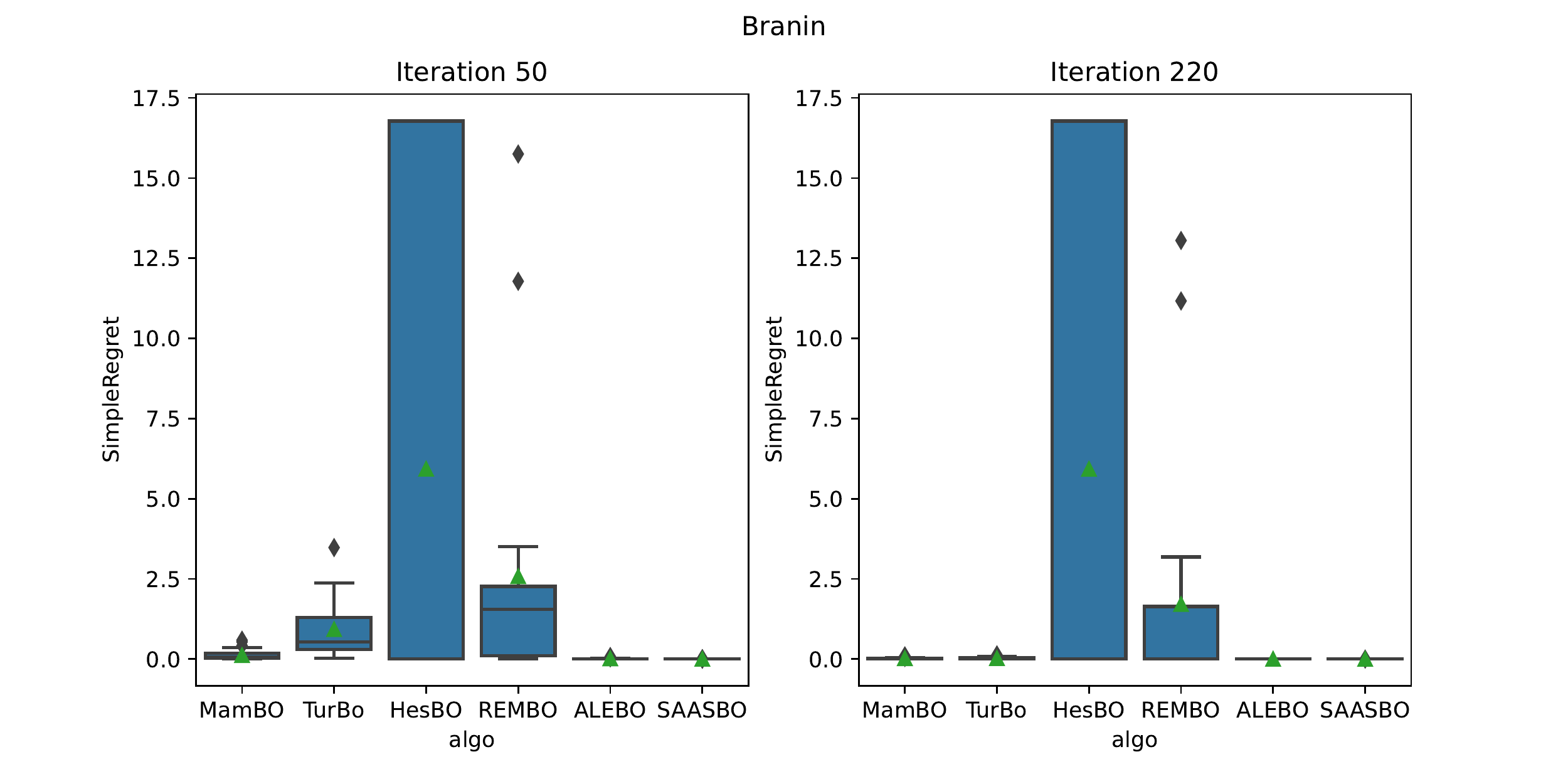}}
\subfigure[Simple regret for 100 dim Camel]{\includegraphics[width=6cm]{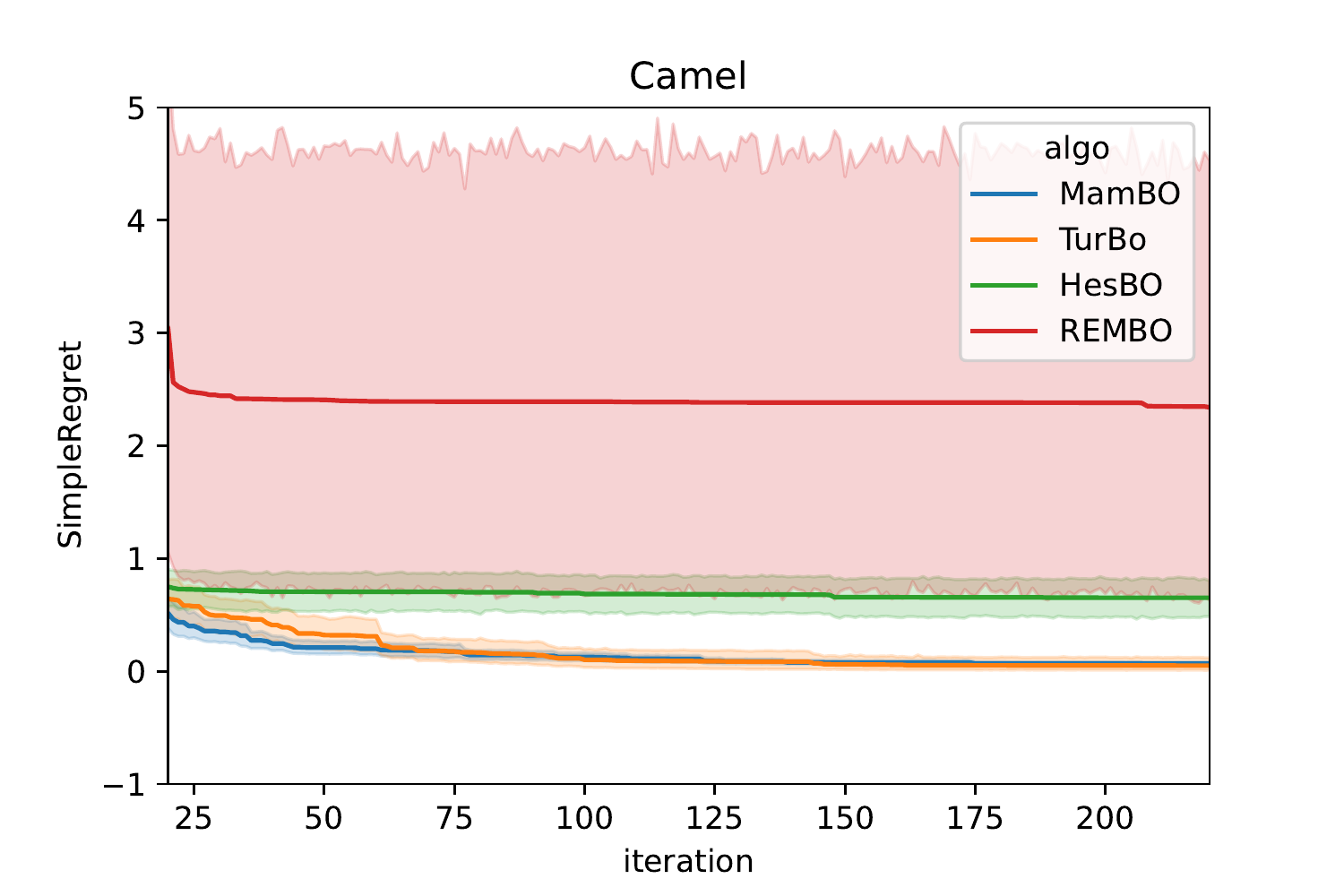}}
\subfigure[Boxplot of simple regret for 100 dim Camel]{\includegraphics[width=8cm]{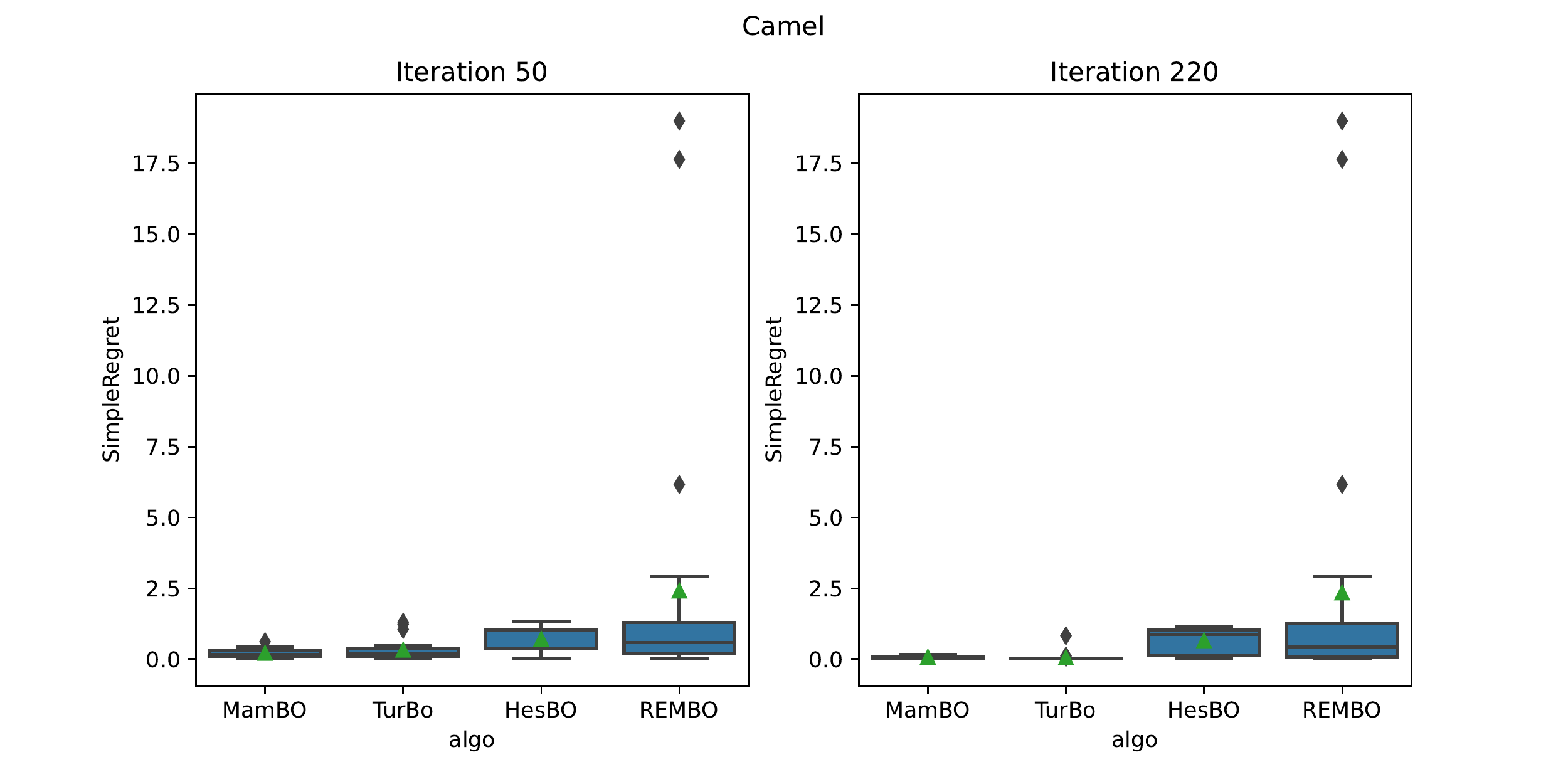}}
\subfigure[Simple regret for 100 dim Eggholder]{\includegraphics[width=6cm]{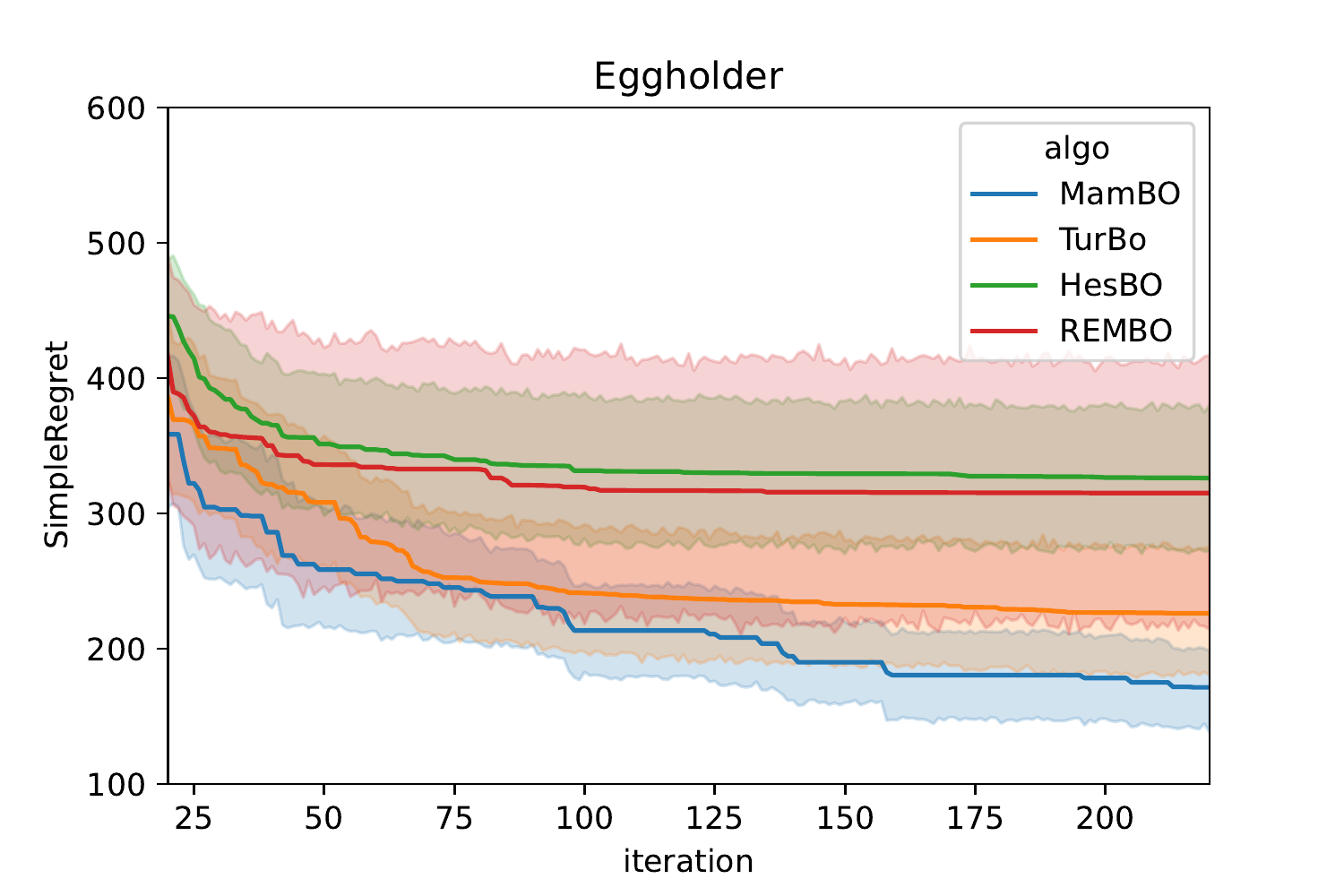}}
\subfigure[Boxplot of simple regret for 100 dim Eggholder]{\includegraphics[width=8cm]{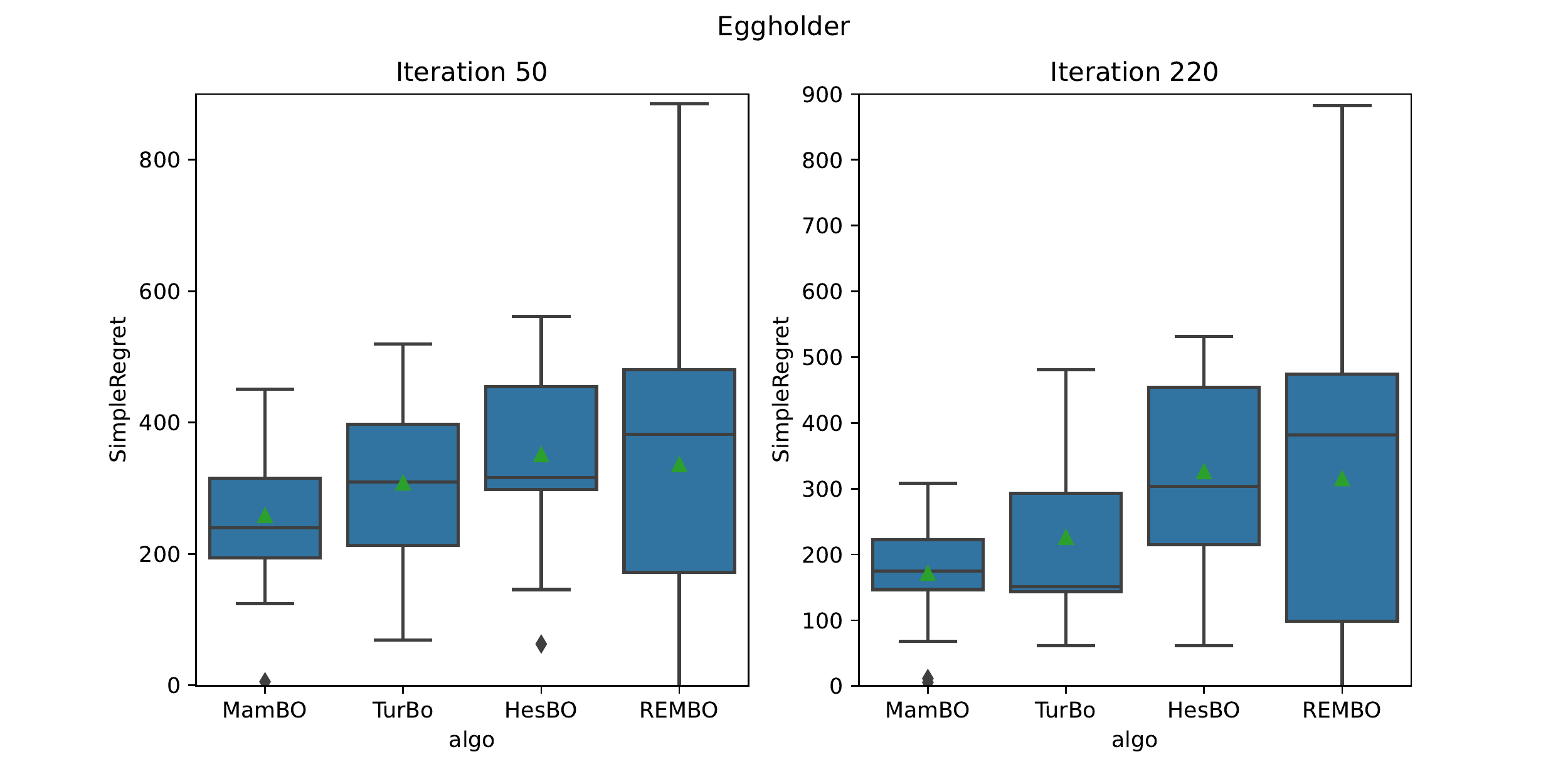}}
\caption{Performance for Branin, Camel and Eggholder. (\textit{Left}) The $x$-axis is the number of iteration, and the $y$-axis is the average simple regret. The 95\% confidence region is provided by 50 macroreplications.  (\textit{Right}) Boxplot of the simple regret after 50 and 220 iterations} 
\label{synthetic}  
\end{figure}

We compare our MamBO with the 5 benchmark algorithms on the following test functions: (1) Branin, (2) Camel, (3) Eggholder with input dimension $d =100$. The exact forms of all test functions are provided in the appendix. For all test functions, 20 initial points are generated, and 200 additional points are selected by the algorithms. We run 50 independent macroreplications in the comparisons. All test functions are set to lie in a 2 dimensional active space. The Branin, Camel and Eggholder functions are all commonly used test functions for optimization, with varying characteristics. The Branin function has three global minima with a flat surface. The Camel function contains 6 local minima, two of which are global. And the Eggholder function has 1 global minimum and is highly multimodal. These test functions can cover a large class of objective functions with different scenarios. In the experiments, the simple regret ($|f(\hat{x}^*)-f(x^*)|$)
is chosen as the performance measure. We consider the heteroscedastic noise in our experiments, and use the Griewank function (divided by the number of active dimensions) as the noise function. The results of the simple regret for all test functions are shown in Figure \ref{synthetic}. 
We use $t$-tests with a significant level of 0.05 to test whether there exist significant differences between the mean simple regret of the different algorithms. From here on, we use the term ``similar performance" to mean that there is no significant difference, and ``outperform" to mean statistically better when comparing two algorithms. We also report the average running time for each algorithm in Table \ref{tab:time}.

\begin{table}[!htb]
\centering
\begin{tabular}{|l|l|l|l|l|l|l|}
\hline
\diagbox{Test problems}{Algorithm} & MamBO & REMBO & HesBO  & TurBO & ALEBO & SAASBO\\ \hline
Branin      & 13.99 & 19.54 & 13.74  & 19.08 & 133.57 & 3186.62\\ \hline
Camel  & 8.61 & 22.09 & 10.38   & 9.00 & - & -\\ \hline
Eggholder   & 6.44 & 24.24 & 9.12  & 7.01 & - & -\\ \hline
Price Optimization   & 8.16 & 10.76 & 7.90 & 7.26 & - & - \\ \hline
VJ          & 68.43 & 68.29 & 103.72 & 180.67& 1121.07 & -\\ \hline
\end{tabular}
\caption{Average running time per iteration in seconds for different problems}
\label{tab:time}
\end{table}

For the Branin function, from Figure \ref{synthetic}(a),  MamBO, ALEBO and SAASBO converge to the true optimum within 50 iterations, whereas TurBO achieves convergence after about 175 iterations. The boxplot (Figure \ref{synthetic}(b)) shows that HesBO and REMBO may converge to a suboptimal or even a wrong optimal point with a wide interquartile range. As the average running time was extremely long for ALEBO and SAASBO, we excluded these two from further test problems.
For the Camel function, from Figure \ref{synthetic}(c), REMBO performs the worst. This poor performance is likely caused by the large information loss due to the large scale random embedding. The confidence interval of the simple regret for REMBO is also very wide, indicating that variability of the solution from REMBO is also high. Our MamBO outperforms both REMBO and HesBO. The boxplot (Figure \ref{synthetic}(d)) shows that the final performance for TurBO and MamBO are quite similar. 
For the Eggholder function, from Figure \ref{synthetic}(e), all four algorithms are stuck at some suboptimal points, and there still remains a gap between the returned value and the true optimal function value. However we can still see a downward trend for MamBO, which indicates its potential for convergence. The boxplot (Figure \ref{synthetic}(f)) shows that MamBO achieves a lowest simple regret and a shortest interquartile range among the four algorithms. From these results, we can see that MamBO can achieve a performance better or comparable to that of other commonly used algorithms.
From Table \ref{tab:time}, we further observe that the computational time required for MamBO is generally less than or at least comparable to that of the other algorithms.

From the results of the above test functions, we see that no single algorithm consistently performs the best. The performance of an algorithm depends on its functional properties. For a flat function, such as the Branin function, each submodel $\M_i$ in MamBO can approximate the true response surface well, and their aggregation reinforces and improves the prediction. From Figure \ref{synthetic}(b) we see that even in the low budget case (after 50 iterations), MamBO can still return a relatively good result. 
For a function with moderate number of local minimum like the Camel function, MamBO and TurBO can find the true optimal if we have sufficient budget. Also, from Figure \ref{synthetic}(d) we clearly see that for other embedding based methods including HesBO and REMBO, the variability of the returned optimum is quite large (larger interquartile range). This is consistent with Example \ref{motivatingexample}, and the embedding-based methods may return suboptimal points when only finite observations are present due to the embedding uncertainty. Our experimental result shows that MamBO can better account for the embedding uncertainty and provide a better result compared to HesBO and REMBO.
For highly multimodal functions like the Eggholder function, with a limited budget, it is very likely for all algorithms to get stuck at some suboptimal point (see from Figure \ref{synthetic}(f)). In general, a larger amount of budget is required to achieve global convergence when faced with highly multi-modal functions. 
In such cases, embedding-based algorithms can perform poorly when the initial design is limited. This is because the initial design for the optimization procedure is selected only in a limited low-dimensional space, and if the embedding obtained is not sufficient for the construction an accurate GP model in this space, the resulting points selected by BO may not enable sufficient and efficient exploration in the original search space. Consequently, the returned optimum is likely to be inaccurate. For algorithms that on local modeling like TurBO, it naturally is more efficient in a local region, and trades that off with less global search. TurBO adopts a local trust region approach with restarts to induce global search in the local approach. However, when the budget is limited, the trust region restarts are likely insufficient to discover the true global optimum area, resulting in TurBO to also be locally stuck too.

\subsection{A Price Optimization Problem}
In this section, we study a specific type of price optimization problem in revenue management from \cite{van2022price}. The input $p=(p_1,\cdots,p_n)$ is the price for a set of products, and the objective is to find the optimal selling price $p^*$ which maximizes the revenue function $\Pi(p)$. Existing studies have shown that such a price optimization problem is typically challenging because $\Pi(p)$ can be highly non-convex and multimodal, and most existing algorithms lack a theoretical guarantee for such problems. Here, we consider a simplified version of the revenue function $\Pi(p)$ as in \cite{van2022price} with number of products $n=10$. The objective function is defined as 
$$f(p_1,\cdots,p_{10})=-\Pi(p_1,\cdots,p_{10})= -\sum_{i =1}^{n} p_{i} \frac{\exp \left(a_{i}-b_{i} p_{i}\right)}{1+\sum_{j=1}^{n} \exp \left(a_{j}-b_{j} p_{j}\right)},$$ where $$a =[4.42,  2.06, -5.32,  0.61, -4.41,  1.90, -5.96, -6.41, -1.82,  3.60],$$ $$b=[0.0010,0.0024,0.0023,0.0057,0.0065,0.0021,0.0080,0.0056,0.0064, 0.0087],$$
and a minus sign is added to the original revenue function as we consider here a minimization problem.
This function is further embedded into a 100 dimensional space to simulate a high dimensional problem.

\begin{figure}[H]
\centering  
\subfigure{
\includegraphics[width=0.6\textwidth]{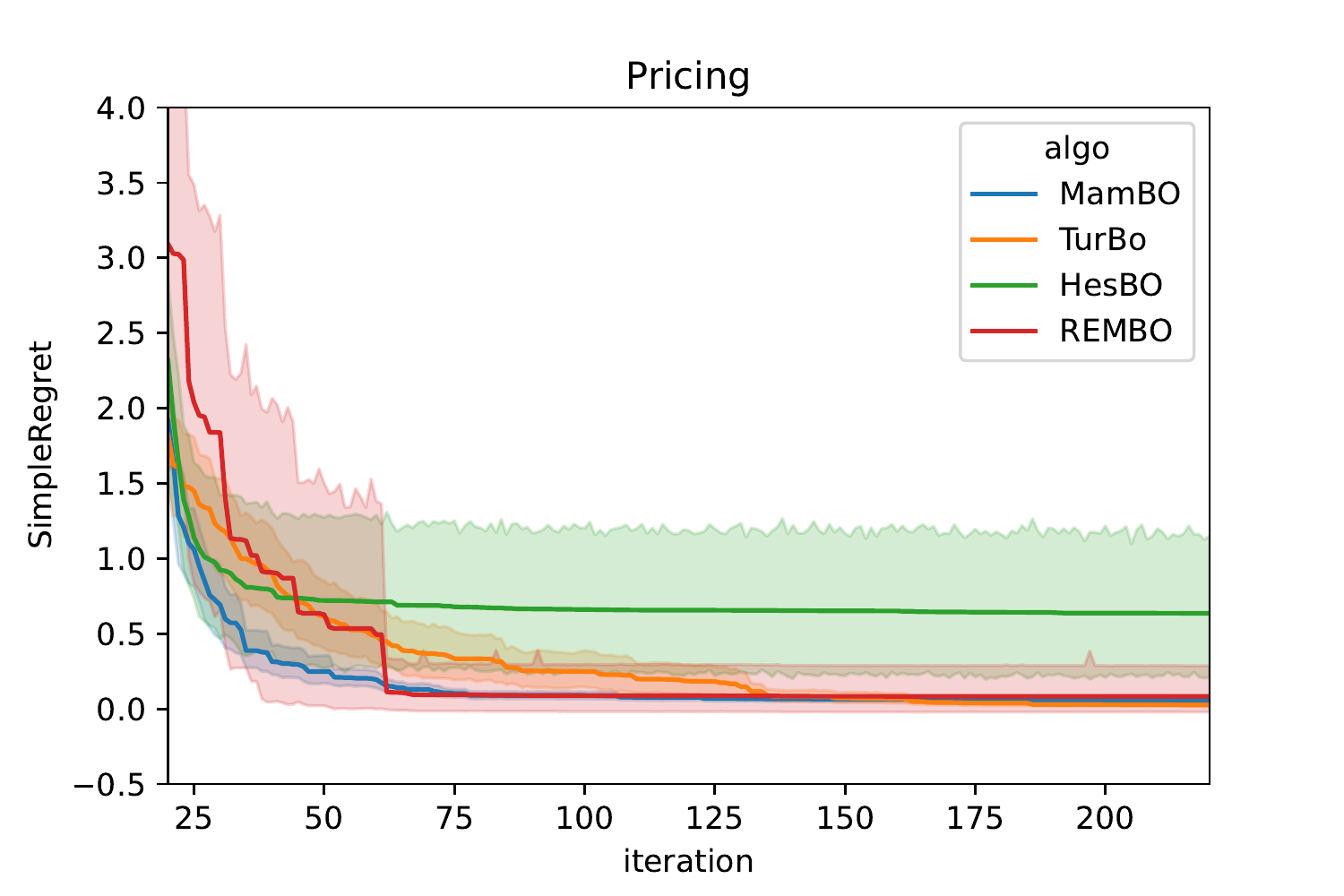}}
\subfigure{
\includegraphics[width=0.6\textwidth]{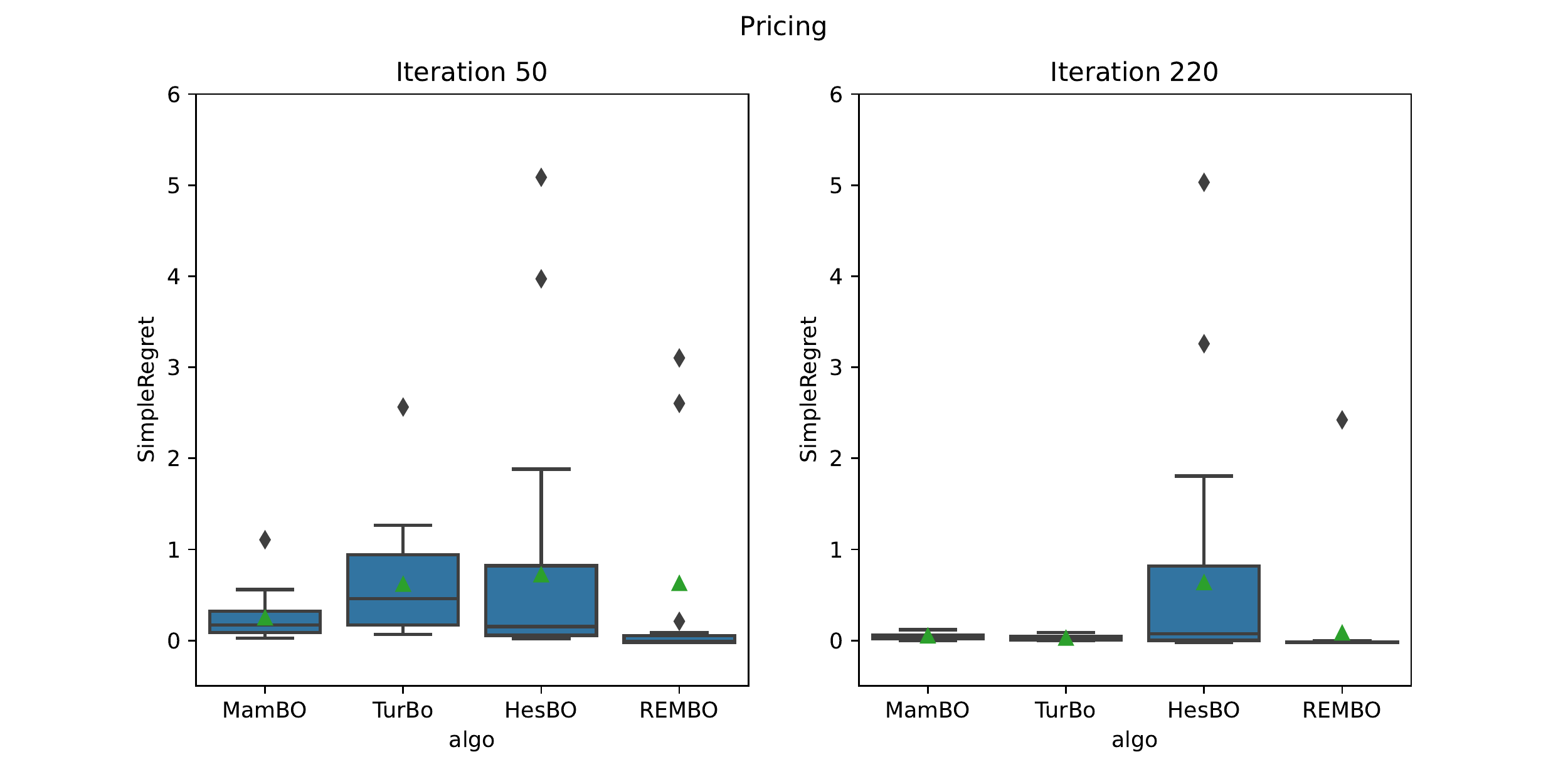}}
\caption{Performance for the price optimization problem. (\textit{Top}) The $x$-axis represents the number of iterations, and the $y$-axis represents the average simple regret. The 95\% confidence region is provided by 50 macroreplications. (\textit{Bottom}) Boxplot of the simple regret after 50 and 220 iterations}
\label{price}
\end{figure}

In this problem, 20 initial points are generated, and 200 additional points are selected by the algorithms. We run 50 independent macroreplications in the comparison. We use the Griewank function
(divided by the number of active dimensions) as the noise function to make the objective function heteroscedastic. Figure \ref{price} shows the performance of the price optimization problem. We choose simple regret as the performance measure.
MamBO and REMBO can find the true optimum after approximately 70 iterations, whereas TurBO can find it after about 130 iterations. HesBO gets trapped in this problem. We also note that, particularly for the starting period, the confidence interval 
of the simple regret for REMBO is much larger than those of the others. Moreover, from the boxplot after 220 iterations, there are more outliers over the upper quartile for HesBO and REMBO, indicating higher variability in the result returned by these algorithms, several of which can be very poor. MamBO's performance on the other hand appear quite robust, likely due to its ability to account for the embedding uncertainty that is ignored by the other algorithms.

\subsection{Viola \& Jones Cascade Classifier (VJ)}\label{sec:vj}
\begin{figure}[H]
\centering  
\subfigure{
\includegraphics[width=0.2\textwidth]{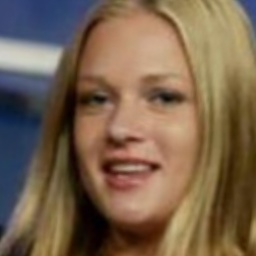}}
\subfigure{
\includegraphics[width=0.2\textwidth]{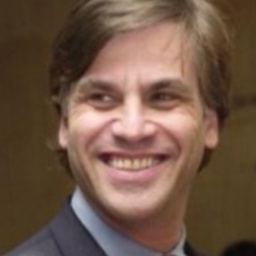}}
\subfigure{
\includegraphics[width=0.2\textwidth]{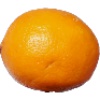}}
\subfigure{
\includegraphics[width=0.2\textwidth]{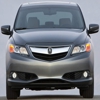}}
\caption{Some faces images and non-faces images in the dataset}
\label{sampleface}
\end{figure}

\begin{figure}[htbp]
\centering  
\subfigure{
\includegraphics[width=0.6\textwidth]{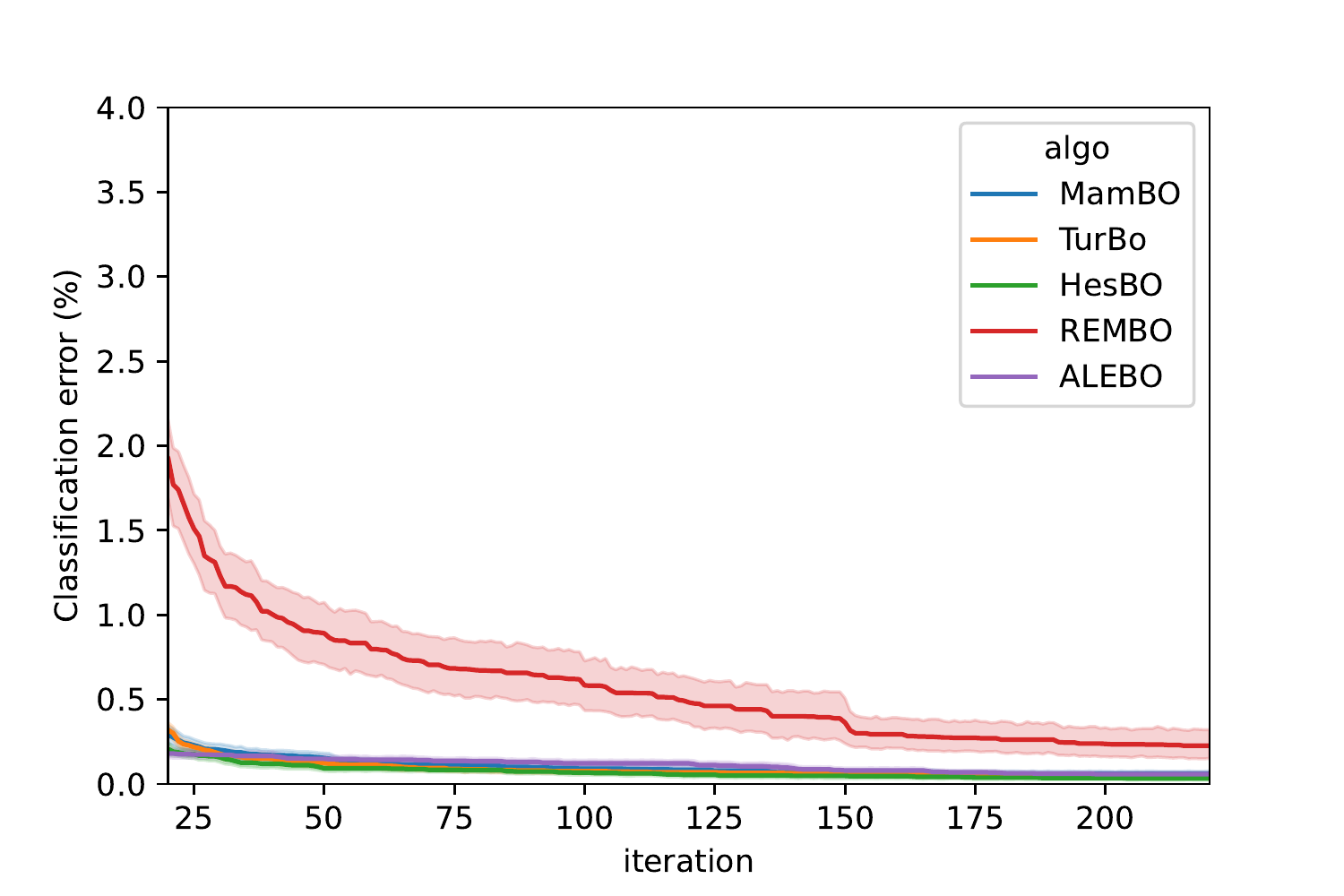}}
\subfigure{
\includegraphics[width=0.6\textwidth]{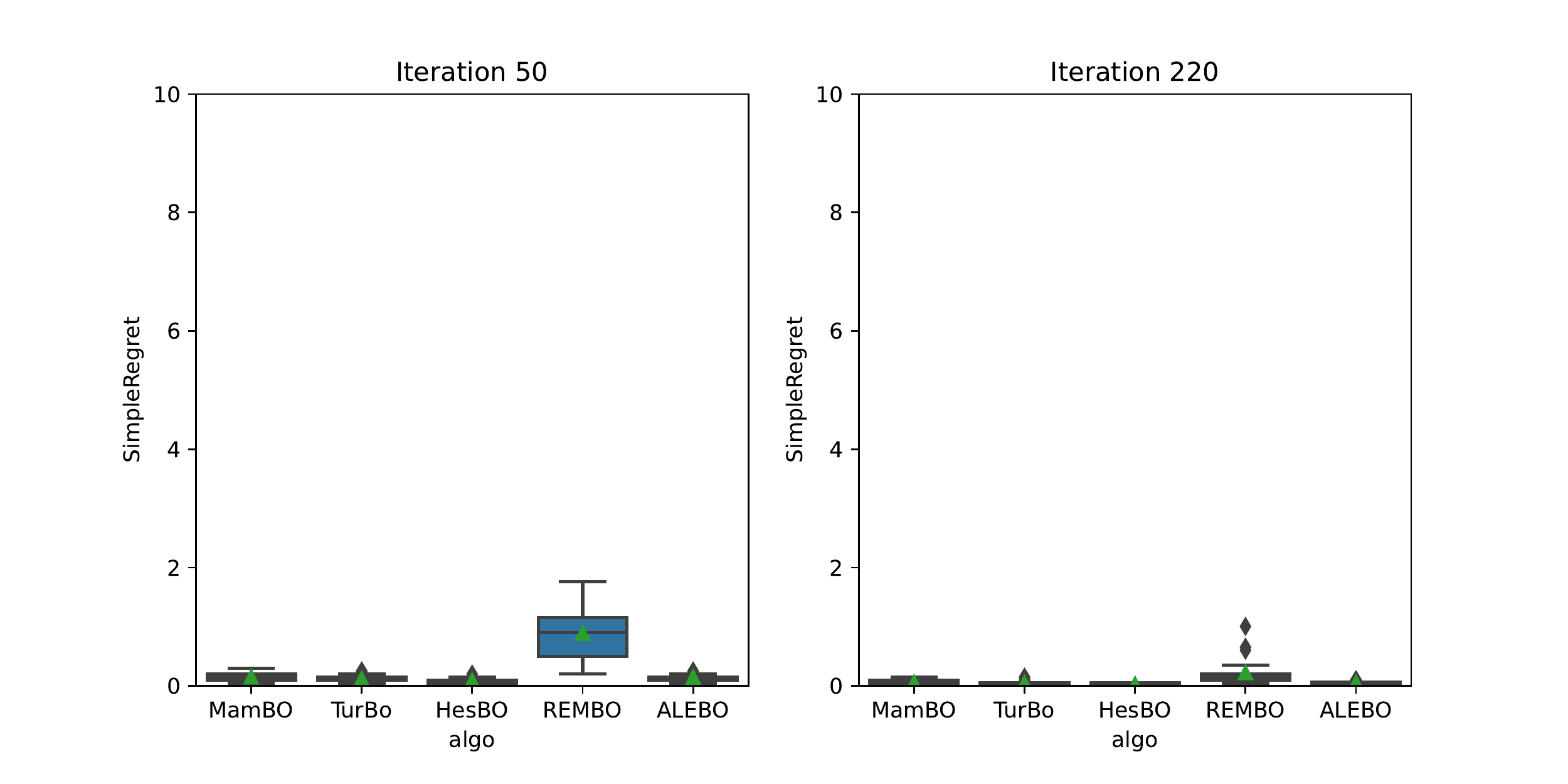}}
\caption{Performance for VJ face detection on the training data. (\textit{Top}) The $x$-axis represents the number of iterations, and the $y$-axis represents the classification error for the training data. The 95\% confidence region is provided by 50 macroreplications. (\textit{Bottom}) Boxplot of the classification error after 50 and 220 iterations}
\label{VJ_train}
\end{figure}

\begin{figure}[htbp]
\centering  
\subfigure{
\includegraphics[width=0.6\textwidth]{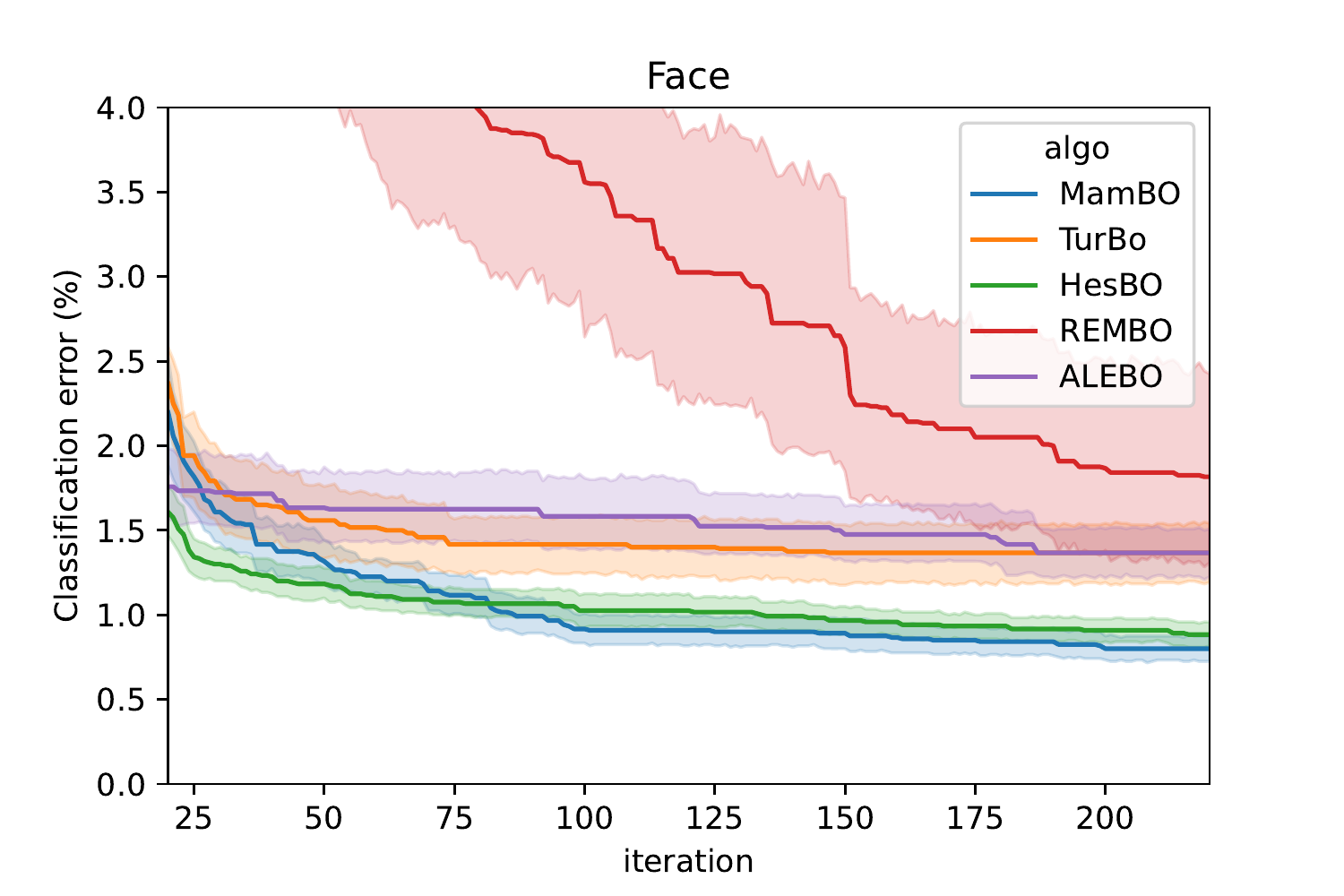}}
\subfigure{
\includegraphics[width=0.6\textwidth]{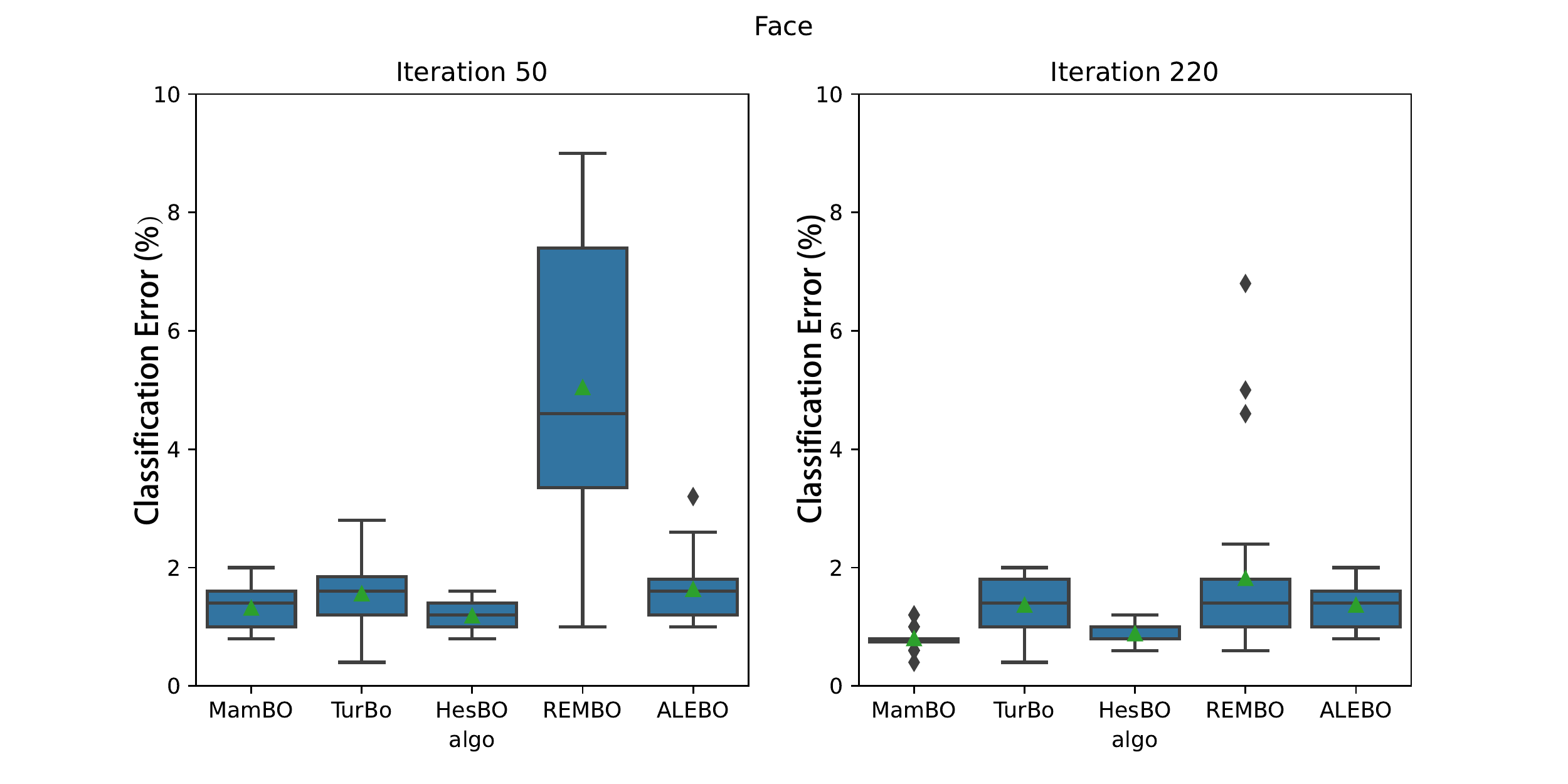}}
\caption{Performance for VJ face detection on the test data. (\textit{Top}) The $x$-axis represents the number of iterations, and the $y$-axis represents the classification error for the test data. The 95\% confidence region is provided by 50 macroreplications. (\textit{Bottom}) Boxplot of the classification error after 50 and 220 iterations}
\label{VJ_test}
\end{figure}

The Viola \& Jones cascade classifier (VJ) \citep{viola2001rapid} is a machine learning algorithm for face detection. The VJ classifier can be used to determine whether an image contains a face. 
The VJ classifier is built using the AdaBoost tree algorithm. To reduce computation time, the VJ classifier employs a cascade system in which  the process of identifying a face is divided into $K=22$ stages. Images that fail to pass in the early stages are immediately classified as “non-face”, which can significantly reduce the computational effort.
Each stage is associated with a threshold, requiring 22 parameters to be optimized. 
Data \footnote{The data are available at  \underline{https://www.kaggle.com/prasunroy/natural-images}.} used in this problem are image files, where some contain a face and others do not.  Examples of the image files are shown in Figure \ref{sampleface}. Here, we focus on minimizing the classification error for the VJ classifier, which requires optimization of 22 parameters. The data are split into training and test datasets in a ratio of 4:1. During the optimization procedure, we randomly select 25\% of the training images to evaluate the classification error for different parameters. Owing to the randomness in the selection of the data used for performance evaluation, the observations are heteroscedastic. To apply the BO algorithms, the GP model is built on the observed classification 
error (in percentage) on the training data and the corresponding parameter values. To solve this problem, 20 initial points are generated (i.e., 20 initial evaluations of the classification error with 20 different sets of parameter values), and 200 additional points are selected by the algorithms. We run 50 independent macroreplications in the comparison. 
In Figure \ref{VJ_train} and Figure \ref{VJ_test}, we report the classification error on the training data and test data as the number of iterations increases.

In Figure \ref{VJ_train}, except for REMBO, the other four algorithms achieve a simple regret near 0 after 50 iterations on the training data, with a reasonably short interquartile range in the boxplots. 

 For real problems however, we are more interested in the algorithms' performance on the test data. On the test data (as shown in Figure \ref{VJ_test}), the performance of MamBO is very similar to that of HesBO, and they outperform the other three algorithms. MamBO achieves a classification error of less than 1\% after 220 iterations on the test data, which is smaller than the accuracy under the default setting provided by the OpenCV package (2.61 \%). Although the performances of MamBO and HesBO are similar, from Table \ref{tab:time} we see that MamBO is almost twice as fast as HesBO. For the VJ problem, the performance for TurBO is not efficient as observed from the test functions. This is probably because the objective for the VJ is quite complicated, and with the limited budget constraint, not all regions can be searched, and hence, TurBO is trapped locally. ALEBO can achieve a relatively good initial performance because of the adoption of the Mahalanobis kernel, which can capture the non-stationarity induced by the embedding, and it provides a better initial fit of the GP model. However, the running time of ALEBO is much longer than the others, which is less desirable when the computational resource is limited. For MamBO, the Bayesian aggregated model provides a better global fit with fewer computational resources. We also see that the interquartile range of the classification error for MamBO is the smallest compared to the others, which highlights the robustness of our proposed algorithm when facing complicated real problems.

In summary, for test functions, MamBO can return reasonably good result if the number of local minimum is small to moderate. It is difficult for all algorithms to achieve convergence to the true optimum with a limited budget when we are facing complicated multimodal functions like the Eggholder function. However, for all test problems, the running time for MamBO was quite fast, and the returned optimum was close to the true optimum. For the practical problems (price optimization and VJ problem), MamBO can still perform effectively and is computationally efficient. In addition, the interquartile range of the simple regret for MamBO was short for all test problems with less or no outliers compared to other embedding-based algorithms. This shows that the Bayesian aggregated model can better account for the embedding uncertainty which in turn can provide a more robust and consistent solution.

\section{Conclusions}\label{conclusion}
In this study, we propose the MamBO algorithm, which is an aggregated model-based Bayesian optimization algorithm for high-dimensional large-scale optimization problems. The algorithm first divides the large-scale data into subsets, and in each subset of data, we fit individual GPs with the subspace embedding technique to deal with high dimensionality. Thereafter, we proceed to our optimization procedure based on an aggregated model that combines the information we have on individual GPs. These techniques can efficiently solve high-dimensional large-scale problems, and also reduce the embedding uncertainty that are largely ignored by other embedding-based BO algorithms. To better address heteroscadestic noise, MamBO introduces an additional allocation stage to allocate computation budgets to sampled points. Furthermore, we prove the convergence of MamBO and study its empirical performance on both synthetic data and real-world computer vision problems. Our numerical results show that MamBO can achieve a comparable performance compared to other commonly used benchmark algorithms.

There are several directions for future work. First, we adopted random embedding in this study. However, for different types of problems, other embedding methods may lead to better performance. It is of interest to further investigate data-adaptive embedding methods. Second, although we apply only the OCBA allocation rule in MamBO, other more sophisticated allocation schemes that satisfy Assumption 1 (which is quite general) can be explored to potentially speed up the convergence of the algorithm. Third, we can further investigate how to train different GP models in parallel to improve the computational efficiency of MamBO further.

\newpage
\bibliography{ref}

\begin{thebibliography}{52}
\providecommand{\natexlab}[1]{#1}
\providecommand{\url}[1]{\texttt{#1}}
\expandafter\ifx\csname urlstyle\endcsname\relax
  \providecommand{\doi}[1]{doi: #1}\else
  \providecommand{\doi}{doi: \begingroup \urlstyle{rm}\Url}\fi

\bibitem[Binois et~al.(2015)Binois, Ginsbourger, and
  Roustant]{binois2015warped}
M.~Binois, D.~Ginsbourger, and O.~Roustant.
\newblock A warped kernel improving robustness in bayesian optimization via
  random embeddings.
\newblock In \emph{International Conference on Learning and Intelligent
  Optimization}, pages 281--286. Springer, 2015.

\bibitem[Burnham and Anderson(2004)]{burnham2004multimodel}
K.~P. Burnham and D.~R. Anderson.
\newblock Multimodel inference: understanding aic and bic in model selection.
\newblock \emph{Sociological methods \& research}, 33\penalty0 (2):\penalty0
  261--304, 2004.

\bibitem[Cartis and Otemissov(2022)]{cartis2022dimensionality}
C.~Cartis and A.~Otemissov.
\newblock A dimensionality reduction technique for unconstrained global
  optimization of functions with low effective dimensionality.
\newblock \emph{Information and Inference: A Journal of the IMA}, 11\penalty0
  (1):\penalty0 167--201, 2022.

\bibitem[Chen et~al.(2000)Chen, Lin, Y{\"u}cesan, and
  Chick]{chen2000simulation}
C.-H. Chen, J.~Lin, E.~Y{\"u}cesan, and S.~E. Chick.
\newblock Simulation budget allocation for further enhancing the efficiency of
  ordinal optimization.
\newblock \emph{Discrete Event Dynamic Systems}, 10\penalty0 (3):\penalty0
  251--270, 2000.

\bibitem[Djolonga et~al.(2013)Djolonga, Krause, and Cevher]{djolonga2013high}
J.~Djolonga, A.~Krause, and V.~Cevher.
\newblock High-dimensional gaussian process bandits.
\newblock \emph{Advances in neural information processing systems}, 26, 2013.

\bibitem[Eriksson and Jankowiak(2021)]{eriksson2021high}
D.~Eriksson and M.~Jankowiak.
\newblock High-dimensional bayesian optimization with sparse axis-aligned
  subspaces.
\newblock In \emph{Uncertainty in Artificial Intelligence}, pages 493--503.
  PMLR, 2021.

\bibitem[Eriksson et~al.(2019)Eriksson, Pearce, Gardner, Turner, and
  Poloczek]{eriksson2019scalable}
D.~Eriksson, M.~Pearce, J.~Gardner, R.~D. Turner, and M.~Poloczek.
\newblock Scalable global optimization via local bayesian optimization.
\newblock \emph{Advances in Neural Information Processing Systems},
  32:\penalty0 5496--5507, 2019.

\bibitem[Fragoso et~al.(2018)Fragoso, Bertoli, and
  Louzada]{fragoso2018bayesian}
T.~M. Fragoso, W.~Bertoli, and F.~Louzada.
\newblock Bayesian model averaging: A systematic review and conceptual
  classification.
\newblock \emph{International Statistical Review}, 86\penalty0 (1):\penalty0
  1--28, 2018.

\bibitem[Frazier et~al.(2009)Frazier, Powell, and
  Dayanik]{frazier2009knowledge}
P.~Frazier, W.~Powell, and S.~Dayanik.
\newblock The knowledge-gradient policy for correlated normal beliefs.
\newblock \emph{INFORMS journal on Computing}, 21\penalty0 (4):\penalty0
  599--613, 2009.

\bibitem[Frazier(2018)]{frazier2018bayesian}
P.~I. Frazier.
\newblock Bayesian optimization.
\newblock In \emph{Recent Advances in Optimization and Modeling of Contemporary
  Problems}, pages 255--278. INFORMS, 2018.

\bibitem[Gonz{\'a}lez et~al.(2016)Gonz{\'a}lez, Dai, Hennig, and
  Lawrence]{gonzalez2016batch}
J.~Gonz{\'a}lez, Z.~Dai, P.~Hennig, and N.~Lawrence.
\newblock Batch bayesian optimization via local penalization.
\newblock In \emph{Artificial intelligence and statistics}, pages 648--657.
  PMLR, 2016.

\bibitem[Guzman et~al.(2020)Guzman, Oliveira, and
  Ramos]{guzman2020heteroscedastic}
R.~Guzman, R.~Oliveira, and F.~Ramos.
\newblock Heteroscedastic bayesian optimisation for stochastic model predictive
  control.
\newblock \emph{IEEE Robotics and Automation Letters}, 6\penalty0 (1):\penalty0
  56--63, 2020.

\bibitem[Hennig and Schuler(2012)]{hennig2012entropy}
P.~Hennig and C.~J. Schuler.
\newblock Entropy search for information-efficient global optimization.
\newblock \emph{Journal of Machine Learning Research}, 13\penalty0 (6), 2012.

\bibitem[Hildebrandt et~al.(2014)Hildebrandt, Goswami, and
  Freitag]{hildebrandt2014large}
T.~Hildebrandt, D.~Goswami, and M.~Freitag.
\newblock Large-scale simulation-based optimization of semiconductor
  dispatching rules.
\newblock In \emph{Proceedings of the Winter Simulation Conference 2014}, pages
  2580--2590. IEEE, 2014.

\bibitem[Hong and Zhang(2021)]{hong2021surrogate}
L.~J. Hong and X.~Zhang.
\newblock Surrogate-based simulation optimization.
\newblock In \emph{Tutorials in Operations Research: Emerging Optimization
  Methods and Modeling Techniques with Applications}, pages 287--311. INFORMS,
  2021.

\bibitem[Jones et~al.(1998)Jones, Schonlau, and Welch]{jones1998efficient}
D.~R. Jones, M.~Schonlau, and W.~J. Welch.
\newblock Efficient global optimization of expensive black-box functions.
\newblock \emph{Journal of Global optimization}, 13\penalty0 (4):\penalty0
  455--492, 1998.

\bibitem[Kandasamy et~al.(2015)Kandasamy, Schneider, and
  P{\'o}czos]{kandasamy2015high}
K.~Kandasamy, J.~Schneider, and B.~P{\'o}czos.
\newblock High dimensional bayesian optimisation and bandits via additive
  models.
\newblock In \emph{International conference on machine learning}, pages
  295--304. PMLR, 2015.

\bibitem[Konishi and Kitagawa(2008)]{konishi2008information}
S.~Konishi and G.~Kitagawa.
\newblock Information criteria and statistical modeling.
\newblock 2008.

\bibitem[Letham et~al.(2020)Letham, Calandra, Rai, and Bakshy]{letham2020re}
B.~Letham, R.~Calandra, A.~Rai, and E.~Bakshy.
\newblock Re-examining linear embeddings for high-dimensional bayesian
  optimization.
\newblock \emph{Advances in neural information processing systems},
  33:\penalty0 1546--1558, 2020.

\bibitem[Li et~al.(2017)Li, Gupta, Rana, Nguyen, Venkatesh, and
  Shilton]{li2017high}
C.~Li, S.~Gupta, S.~Rana, V.~Nguyen, S.~Venkatesh, and A.~Shilton.
\newblock High dimensional bayesian optimization using dropout.
\newblock In \emph{Proceedings of the 26th International Joint Conference on
  Artificial Intelligence}, pages 2096--2102, 2017.

\bibitem[Li et~al.(2016)Li, Kandasamy, P{\'o}czos, and Schneider]{li2016high}
C.-L. Li, K.~Kandasamy, B.~P{\'o}czos, and J.~Schneider.
\newblock High dimensional bayesian optimization via restricted projection
  pursuit models.
\newblock In \emph{Artificial Intelligence and Statistics}, pages 884--892.
  PMLR, 2016.

\bibitem[Locatelli(1997)]{locatelli1997bayesian}
M.~Locatelli.
\newblock Bayesian algorithms for one-dimensional global optimization.
\newblock \emph{Journal of Global Optimization}, 10\penalty0 (1):\penalty0
  57--76, 1997.

\bibitem[Marmin et~al.(2015)Marmin, Chevalier, and
  Ginsbourger]{marmin2015differentiating}
S.~Marmin, C.~Chevalier, and D.~Ginsbourger.
\newblock Differentiating the multipoint expected improvement for optimal batch
  design.
\newblock In \emph{International Workshop on Machine Learning, Optimization and
  Big Data}, pages 37--48. Springer, 2015.

\bibitem[McIntire et~al.(2016)McIntire, Ratner, and Ermon]{mcintire2016sparse}
M.~McIntire, D.~Ratner, and S.~Ermon.
\newblock Sparse gaussian processes for bayesian optimization.
\newblock In \emph{UAI}, 2016.

\bibitem[Meng et~al.(2022)Meng, Wang, and Ng]{meng2022combined}
Q.~Meng, S.~Wang, and S.~H. Ng.
\newblock Combined global and local search for optimization with gaussian
  process models.
\newblock \emph{INFORMS Journal on Computing}, 34\penalty0 (1):\penalty0
  622--637, 2022.

\bibitem[Nayebi et~al.(2019)Nayebi, Munteanu, and
  Poloczek]{nayebi2019framework}
A.~Nayebi, A.~Munteanu, and M.~Poloczek.
\newblock A framework for bayesian optimization in embedded subspaces.
\newblock In \emph{International Conference on Machine Learning}, pages
  4752--4761. PMLR, 2019.

\bibitem[Ng and Yin(2012)]{ng2012bayesian}
S.~H. Ng and J.~Yin.
\newblock Bayesian kriging analysis and design for stochastic simulations.
\newblock \emph{ACM Transactions on Modeling and Computer Simulation (TOMACS)},
  22\penalty0 (3):\penalty0 1--26, 2012.

\bibitem[Nickson et~al.(2014)Nickson, Osborne, Reece, and
  Roberts]{nickson2014automated}
T.~Nickson, M.~A. Osborne, S.~Reece, and S.~Roberts.
\newblock Automated machine learning using stochastic algorithm tuning.
\newblock In \emph{NIPS Workshop on Bayesian Optimization}, 2014.

\bibitem[Oh et~al.(2018)Oh, Gavves, and Welling]{oh2018bock}
C.~Oh, E.~Gavves, and M.~Welling.
\newblock Bock: Bayesian optimization with cylindrical kernels.
\newblock In \emph{International Conference on Machine Learning}, pages
  3868--3877. PMLR, 2018.

\bibitem[Paul et~al.(2014)Paul, Boutsidis, Magdon-Ismail, and
  Drineas]{paul2014random}
S.~Paul, C.~Boutsidis, M.~Magdon-Ismail, and P.~Drineas.
\newblock Random projections for linear support vector machines.
\newblock \emph{ACM Transactions on Knowledge Discovery from Data (TKDD)},
  8\penalty0 (4):\penalty0 1--25, 2014.

\bibitem[Pedrielli et~al.(2020)Pedrielli, Wang, and Ng]{pedrielli2020extended}
G.~Pedrielli, S.~Wang, and S.~H. Ng.
\newblock An extended two-stage sequential optimization approach: Properties
  and performance.
\newblock \emph{European Journal of Operational Research}, 287\penalty0
  (3):\penalty0 929--945, 2020.

\bibitem[Phillips(2021)]{phillips2021pricing}
R.~L. Phillips.
\newblock \emph{Pricing and revenue optimization}.
\newblock Stanford university press, 2021.

\bibitem[Quan et~al.(2013)Quan, Yin, Ng, and Lee]{quan2013simulation}
N.~Quan, J.~Yin, S.~H. Ng, and L.~H. Lee.
\newblock Simulation optimization via kriging: a sequential search using
  expected improvement with computing budget constraints.
\newblock \emph{Iie Transactions}, 45\penalty0 (7):\penalty0 763--780, 2013.

\bibitem[Russo et~al.(2018)Russo, Van~Roy, Kazerouni, Osband, Wen,
  et~al.]{russo2018tutorial}
D.~J. Russo, B.~Van~Roy, A.~Kazerouni, I.~Osband, Z.~Wen, et~al.
\newblock A tutorial on thompson sampling.
\newblock \emph{Foundations and Trends{\textregistered} in Machine Learning},
  11\penalty0 (1):\penalty0 1--96, 2018.

\bibitem[Snoek et~al.(2012)Snoek, Larochelle, and Adams]{snoek2012practical}
J.~Snoek, H.~Larochelle, and R.~P. Adams.
\newblock Practical bayesian optimization of machine learning algorithms.
\newblock \emph{Advances in neural information processing systems}, 25, 2012.

\bibitem[Snoek et~al.(2015)Snoek, Rippel, Swersky, Kiros, Satish, Sundaram,
  Patwary, Prabhat, and Adams]{snoek2015scalable}
J.~Snoek, O.~Rippel, K.~Swersky, R.~Kiros, N.~Satish, N.~Sundaram, M.~Patwary,
  M.~Prabhat, and R.~Adams.
\newblock Scalable bayesian optimization using deep neural networks.
\newblock In \emph{International conference on machine learning}, pages
  2171--2180. PMLR, 2015.

\bibitem[Srinivas et~al.(2010)Srinivas, Krause, Kakade, and
  Seeger]{Srinivas2010GaussianPO}
N.~Srinivas, A.~Krause, S.~Kakade, and M.~Seeger.
\newblock Gaussian process optimization in the bandit setting: No regret and
  experimental design.
\newblock In \emph{International Conference on Machine Learning}, page
  1015–1022, 2010.

\bibitem[Stuart and Teckentrup(2018)]{stuart2018posterior}
A.~Stuart and A.~Teckentrup.
\newblock Posterior consistency for gaussian process approximations of bayesian
  posterior distributions.
\newblock \emph{Mathematics of Computation}, 87\penalty0 (310):\penalty0
  721--753, 2018.

\bibitem[Tanaka et~al.(2019)Tanaka, Tanaka, Iwata, Kurashima, Okawa, Akagi, and
  Toda]{tanaka2019spatially}
Y.~Tanaka, T.~Tanaka, T.~Iwata, T.~Kurashima, M.~Okawa, Y.~Akagi, and H.~Toda.
\newblock Spatially aggregated gaussian processes with multivariate areal
  outputs.
\newblock In \emph{Proceedings of the 33rd International Conference on Neural
  Information Processing Systems}, pages 3005--3015, 2019.

\bibitem[Thompson(1933)]{thompson1933likelihood}
W.~R. Thompson.
\newblock On the likelihood that one unknown probability exceeds another in
  view of the evidence of two samples.
\newblock \emph{Biometrika}, 25\penalty0 (3/4):\penalty0 285--294, 1933.

\bibitem[van~de Geer and den Boer(2022)]{van2022price}
R.~van~de Geer and A.~V. den Boer.
\newblock Price optimization under the finite-mixture logit model.
\newblock \emph{Management Science}, 2022.

\bibitem[Viola and Jones(2001)]{viola2001rapid}
P.~Viola and M.~Jones.
\newblock Rapid object detection using a boosted cascade of simple features.
\newblock In \emph{Proceedings of the 2001 IEEE computer society conference on
  computer vision and pattern recognition. CVPR 2001}, volume~1, pages I--I.
  Ieee, 2001.

\bibitem[Wang et~al.(2020)Wang, Clark, Liu, and Frazier]{wang2020parallel}
J.~Wang, S.~C. Clark, E.~Liu, and P.~I. Frazier.
\newblock Parallel bayesian global optimization of expensive functions.
\newblock \emph{Operations Research}, 68\penalty0 (6):\penalty0 1850--1865,
  2020.

\bibitem[Wang et~al.(2021)Wang, Ng, and Haskell]{wang2021multilevel}
S.~Wang, S.~H. Ng, and W.~B. Haskell.
\newblock A multilevel simulation optimization approach for quantile functions.
\newblock \emph{INFORMS Journal on Computing}, 2021.

\bibitem[Wang et~al.(2016)Wang, Hutter, Zoghi, Matheson, and
  de~Feitas]{wang2016bayesian}
Z.~Wang, F.~Hutter, M.~Zoghi, D.~Matheson, and N.~de~Feitas.
\newblock Bayesian optimization in a billion dimensions via random embeddings.
\newblock \emph{Journal of Artificial Intelligence Research}, 55:\penalty0
  361--387, 2016.

\bibitem[Wang et~al.(2018)Wang, Gehring, Kohli, and Jegelka]{wang2018batched}
Z.~Wang, C.~Gehring, P.~Kohli, and S.~Jegelka.
\newblock Batched large-scale bayesian optimization in high-dimensional spaces.
\newblock In \emph{International Conference on Artificial Intelligence and
  Statistics}, pages 745--754. PMLR, 2018.

\bibitem[Wasserman(2000)]{wasserman2000bayesian}
L.~Wasserman.
\newblock Bayesian model selection and model averaging.
\newblock \emph{Journal of mathematical psychology}, 44\penalty0 (1):\penalty0
  92--107, 2000.

\bibitem[Williams and Rasmussen(2006)]{williams2006gaussian}
C.~K. Williams and C.~E. Rasmussen.
\newblock \emph{Gaussian processes for machine learning}, volume~2.
\newblock MIT press Cambridge, MA, 2006.

\bibitem[Xie et~al.(2014)Xie, Nelson, and Barton]{xie2014bayesian}
W.~Xie, B.~L. Nelson, and R.~R. Barton.
\newblock A bayesian framework for quantifying uncertainty in stochastic
  simulation.
\newblock \emph{Operations Research}, 62\penalty0 (6):\penalty0 1439--1452,
  2014.

\bibitem[Xuereb et~al.(2020)Xuereb, Ng, and Pedrielli]{xuereb2020stochastic}
M.~Xuereb, S.~H. Ng, and G.~Pedrielli.
\newblock Stochastic gaussian process model averaging for high-dimensional
  inputs.
\newblock In \emph{2020 Winter Simulation Conference (WSC)}, pages 373--384.
  IEEE, 2020.

\bibitem[Yu and Gen(2010)]{yu2010introduction}
X.~Yu and M.~Gen.
\newblock \emph{Introduction to evolutionary algorithms}.
\newblock Springer Science \& Business Media, 2010.

\bibitem[Zhigljavsky(2012)]{zhigljavsky2012theory}
A.~A. Zhigljavsky.
\newblock \emph{Theory of global random search}, volume~65.
\newblock Springer Science \& Business Media, 2012.

\end{thebibliography}

\newpage
\begin{appendix}
\section{Proof of Theorem \ref{thm:bound}}
For each submodel $\M_i\sim\mathcal{GP}(m_i,C_i)$, we have
$\mu_n=\sum_{i=1}^m w_im_i$.
Since $\left|\mu_n-f(x)\right|\leq \left|\mu_n-\sum_{i=1}^m w_i\M_i\right|+\left|\sum_{i=1}^m w_i\M_i-f(x)\right|$ from triangular inequality, for all $\epsilon>0$, we have
\begin{align}\label{eq: decompose}
\P\left(\left|\mu_n-f(x)\right|\geq 2\epsilon\right)&\leq
\P\left(\left|\mu_n-\sum_{i=1}^m w_i\M_i\right|+\left|\sum_{i=1}^m w_i\left(\M_i-f(x)\right)\right|\geq 2\epsilon\right)\\
&\leq \P\left(\left|\sum_{i=1}^m w_i\left(m_i-\M_i\right)\right|\geq \epsilon\right) + \P\left(\left|\sum_{i=1}^m w_i\left(\M_i-f(x)\right)\right|\geq \epsilon\right).
\end{align}

By Chebyshev’s Inequality and Propositions
3.3, 3.4 and 3.5 of \cite{stuart2018posterior} we have for all $i$
\[
\P\left(\left|m_i-\M_i\right|\geq \epsilon \right)\leq  \frac{C_i(x,x)}{\epsilon^2} \rightarrow 0 \quad\text{uniformly in }x,
\]
i.e. $\left|m_i-\M_i\right|\rightarrow_p 0$ uniformly in $x$.
Hence, 
\[\sup_x\left|\sum_{i=1}^mw_i(m_i-\M_i)\right|\leq m\sup_{i}\sup_x\left|w_i(m_i-\M_i)\right|\rightarrow_p 0 \quad \text{for fixed }m,\]
i.e. $\left|\sum_{i=1}^mw_i(m_i-\M_i)\right|\rightarrow_p 0$ uniformly in $x$.

Based on Theorem 2 and Corollary 8 of \cite{nayebi2019framework}, there exists a constant $B$ such that $|m_i(x)-f(x)|\leq 5\epsilon' \norm{x}B$, where $\epsilon'\in (0,\frac12)$ is an approximation parameter, which reflects the information lose for each subspace embedding. And in general, the constant $B$ depends on the choice of the kernel function.

Hence,
\[
\begin{aligned}
&\P\left(\left|\M_i - f(x)\right|\geq 10\epsilon' \norm{x}B\right)\\ &\leq\P\left(\left|\M_i - m_i(x)\right|\geq 5\epsilon' \norm{x}B\right) + \P\left(\left|m_i(x) - f(x)\right|\geq 5\epsilon' \norm{x}B\right)\\
&\leq \frac{C_i(x,x)}{25\epsilon'^2\norm{x}^2B^2}+0\\
&\rightarrow 0.
\end{aligned}
\]
Therefore, based on triangle inequality, it is not difficult to see
\[
\begin{aligned}
\P\left(\left|\sum_{i=1}^m w_i(\M_i-f(x))\right|\geq 10 \epsilon' \norm{x}B\right) &\leq \P\left(\sum_{i=1}^m w_i\left|(\M_i-f(x))\right|\geq 10 \epsilon' \norm{x}B\right)\\
&\leq \sum_{i=1}^m \P\left(w_i\left|(\M_i-f(x))\right|\geq 10 \epsilon' \norm{x}B\right)\\
&\leq \sum_{i=1}^m \P\left(\left|(\M_i-f(x))\right|\geq 10 \epsilon' \norm{x}B\right)\\
&\rightarrow 0 \quad \text{for fixed }m.
\end{aligned}
\]
In view of the above display and Equation \eqref{eq: decompose} and taking $\epsilon=10\epsilon'\norm{x}B$, we conclude
\[\P\left(\left|\mu_n-f(x)\right|\geq 20\epsilon'\norm{x}B\right)\rightarrow 0 \quad \text{as }n \rightarrow \infty.\]
Replacing $20B$ by a new constant, we get exactly the same form as in Theorem \ref{thm:bound}.

\section{Proof of Lemma 1}
In this section, we will prove that the points visited by the algorithm is dense. The proof is divided into three steps. First, we will construct an upper bound of $\mathrm{EI}_T(x)$ for any unsampled points. Second, we will construct a region around any design point where $\mathrm{EI}_T(x)$ is bounded above by a threshold $c$. Third, we will conclude the dense result for the visited points.

\subsection{Upper bound for $\mathrm{EI}_T(x)$}
Recall that in iteration $n-n_0$ \[
\mathrm{EI}_{T}\left(x\right)
=\Delta \Phi\left(\frac{\Delta}{\sqrt{k_{n}\left(x,x\right)}}\right)+\sqrt{k_{n}\left(x,x\right)} \phi\left(\frac{\Delta}{\sqrt{k_{n}\left(x,x\right)}}\right)
,\]
where $T=\min\{\bar{y}(x_1),\cdots,\bar{y}(x_n)\},\Delta=T-\mu_{n}(x),k_n(x,x)=\sum_{i=1}^m w_i^2 C_i(x,x)$.
Without loss of generality, suppose $\norm{x}=1$. Based on Assumption 2 and Theorem \ref{thm:bound}, there exists a large value $M$ such that for large enough $n$ the noisy observation $\bar{y}(x)$ is bounded in $(-M,M)$, and predictor $\mu_n(x)$ is bounded in $(-M-\epsilon'B,M+\epsilon'B)$ for all $x$. Therefore,
\begin{equation}
    -2M-\epsilon'B\leq \Delta \leq 2M+\epsilon'B
\end{equation}

As $\mathrm{EI}_T(x)$ is an increasing function of $\Delta$, we have 
\begin{equation}
    \mathrm{EI}_{T}\left(x\right)
\leq(2M+\epsilon'B) \Phi\left(\frac{2M+\epsilon'B}{\sqrt{k_{n}\left(x,x\right)}}\right)+\sqrt{k_{n}\left(x,x\right)} \phi\left(\frac{2M+\epsilon'B}{\sqrt{k_{n}\left(x,x\right)}}\right):=P_n(x)
\end{equation}
Let $x_0$ be the design point with the largest correlation with $x$. We next aim to get an upper bound for $k_n(x,x)$. Denote $\Sigma = \Sigma_F+\Sigma_\xi=\left[\begin{array}{ll}
\Sigma_{11} & \Sigma_{12} \\
\Sigma_{21} & \Sigma_{22}
\end{array}\right]$, where $\Sigma_{11}$ is the variance at $x_0$, $\Sigma_{22}$ is the the variance-covariance matrix at all other design point except for $x_0$, and $\Sigma_{21}=\Sigma_{12}^T$ is the covariance between $x_0$ and other design points. Further denote $\Sigma_F(x,\cdot)=(\sigma_F^2\mathrm{corr}(x,x_0),t(x)^T)^T$.
For large enough $n$, we have
\begin{equation}
\begin{aligned}
C_i(x,x) &= \sigma_F^2-\Sigma_F(x,\cdot)^T\Sigma^{-1}\Sigma_F(x,\cdot)+u(x)^T\left[\Omega^{-1}+L\Sigma^{-1}L^T\right]^{-1}u(x)\\
&=\sigma_F^2-\frac{\sigma_F^{4} \operatorname{corr}\left(x, x_{0}\right)^{2}}{\Sigma_{11}}-\Gamma^{T} P \Gamma +u(x)^T\left[\Omega^{-1}+L\Sigma^{-1}L^T\right]^{-1}u(x)\\
&\leq \sigma_F^2-\sigma_F^2 \operatorname{corr}\left(x, x_{0}\right)^{2} + u(x)^T\left[\Omega^{-1}+L\Sigma^{-1}L^T\right]^{-1}u(x),
\end{aligned}
\end{equation}
where $\Gamma=\Sigma_{21} \Sigma_{11}^{-1} \sigma^{2} \operatorname{corr}\left(x, x_{0}\right)-t(x) \text { and } P^{-1}=\Sigma_{22}-\Sigma_{21} \Sigma_{11}^{-1} \Sigma_{12}$.
For the last term, recall that $u(x)=l(x)-L\Sigma^{-1}\Sigma_F(x,\cdot)$. For the commonly used $l(x)=1$, $L\Sigma^{-1}\Sigma_F(x,\cdot)$ is the GP prediction given $L$. According to the argument in \cite{wang2021multilevel}, we have $|u(x)|=\mathcal{O}(|x-x_0|)$. Hence, there exists $M_1$ such that $|u(x)|<M_1(|x-x_0|)$. Also, $L\Sigma^{-1}L^T>\frac{1}{\sigma_F^2+\sigma_0^2}$, so we have
$u(x)^T\left[\Omega^{-1}+L\Sigma^{-1}L^T\right]^{-1}u(x)=|u(x)|^2\left[\Omega^{-1}+L\Sigma^{-1}L^T\right]^{-1}<M_1^2|x-x_0|^2\left[\Omega^{-1}+(\sigma_F^2+\sigma_0^2)^{-1}\right]:=M_2|x-x_0|^2$.
Hence,
\begin{equation}
    C_i(x,x)\leq\sigma_F^2-\sigma_F^2 \operatorname{corr}\left(x, x_{0}\right)^{2} +M_2|x-x_0|^2
\end{equation}

And
\begin{equation}
\begin{aligned}
k_n(x,x)&=\sum_{i=1}^mw_i^2C_i(x,x)\\
&\leq m \sup_i C_i(x,x)\\
&\leq m \left\{\sigma_F^2-\sigma_F^2 \operatorname{corr}\left(x, x_{0}\right)^{2} +M_2|x-x_0|^2\right\}\\
&:=mk_0(x;x_0)
\end{aligned}
\end{equation}
We can check that $k_0(x;x_0)$ is an increasing function of $|x-x_0|$, so we have
\begin{equation}
\mathrm{EI}_T(x)\leq P_n(x)\leq(2M+\epsilon'B) \Phi\left(\frac{2M+\epsilon'B}{\sqrt{mk_0(x;x_0)}}\right)+\sqrt{mk_0(x;x_0)} \phi\left(\frac{2M+\epsilon'B}{\sqrt{mk_0(x;x_0)}}\right):=G_n(x;x_0)
\end{equation}
We can check that $G_n(x;x_0)$ is also an increasing function of $|x-x_0|$.

\subsection{Region construction}
For any design point $x_0$, we can construct
$$R\left(x_{0}, c\right)=\left\{x \in \mathcal{X} \mid G_{n}\left(x ; x_{0}\right)<c\right\}$$
which contains $x_0$. Since $G_n(x;x_0)$ is a monotonic increasing function of $|x-x_0|$ and it will tends to 0 as $x \to x_0$.
Hence, $R(x_0,c)$ is a region centered at $x_0$, such that for all $x$ in this region, $\mathrm{EI}_T(x) <c$.

\subsection{Proof of density}
Consider the stopping rule that stops the algorithm when EI$_T(x_{n+1})\leq c$, where $c$ is a pre-specified threshold. With this stopping rule, the points in $R(x_0,c)$ will never be selected. However, if we let the total budget tends to infinity, then MamBO will never stop in finite number of iterations. To achieve this, we will decrease $c$ once $\mathrm{EI}_T(x)<c$. By Theorem 1 and Lemma 1 in \cite{locatelli1997bayesian}, we can show that the visited points will eventually be dense over the decision space if $c$ keeps decreasing to zero.

\section{Proof of Theorem \ref{thm:conv}}
In iteration $t$, let $N_t$ be the total number of design points, and $D_t$ be the whole design points set, $\hat{x}_t$ be the current best such that $\bar{y}_t^*=\bar{y}_t(\hat{x}_t)=\min _{x \in D_t}\bar{y}_t(x)$, and $x_0^t$ is the true best in $D_t$ such that $f(x_0^t)=\min_{x \in D_t}f(x)$. We aim to prove that $\bar{y}_t^* \to f(x^*)$ w.p.1 as $t \to \infty$. This can be divided into three steps. First, we prove that $\bar{y}_t^* - f(x_0^t) \to 0$ w.p.1, then prove $ f(x_0^t)-f(x^*)$ w.p.1. Then we can conclude the result.

\subsection{$\bar{y}_t^* - f(x_0^t) \to 0$ w.p.1 as $t \to \infty$}

The sufficient condition is that $\sum_{t=1}^{\infty} \mathbb{P}\left[\left|\bar{y}_{t}^{*}-f\left(x_{0}^{t}\right)\right|>\delta\right]<\infty$, for any $\delta>0 .$ Note that
\begin{equation}
\begin{aligned}
\mathbb{P}\left[\left|\bar{y}_{t}^{*}-f\left(x_{0}^{t}\right)\right|>\delta\right] &=\mathbb{P}\left[\left|\bar{y}_{t}\left(\hat{x}_{t}\right)-f\left(\hat{x}_{t}\right)+f\left(\hat{x}_{t}\right)-f\left(x_{0}^{t}\right)\right|>\delta\right] \\
&<\underbrace{\mathbb{P}\left[\left|\bar{y}_{t}\left(\hat{x}_{t}\right)-f\left(\hat{x}_{t}\right)\right|>\frac{\delta}{2}\right]}_{(\romannumeral1)}+\underbrace{\mathbb{P}\left[\left|f\left(\hat{x}_{t}\right)-f\left(x_{0}^{t}\right)\right|>\frac{\delta}{2}\right]}_{(\romannumeral2)}
\end{aligned}
\end{equation}
We first bound the term $(\romannumeral1)$. Define $\sigma_{0}^{2}:=\max _{x \in \mathcal{X}} \sigma_{\xi}^{2}(x) .$ 

For any $x \in D_{t}$, $\bar{y}_{t}(x)-f(x) \sim \mathcal{N}\left(0, \sigma_{\epsilon}^{2}(x) / M_{t}(x)\right)$, and thus by Gaussian tail inequality,
\begin{equation}
\mathbb{P}\left[\left|\bar{y}_{t}(x)-f(x)\right|>\frac{\delta}{2}\right] \leq 2 \exp \left(-\frac{\delta^{2} M_{t}(x)}{8 \sigma_{\xi}^{2}(x)}\right) \leq 2 \exp \left(-\frac{\delta^{2} s_{N_{t}}}{8 \sigma_{0}^{2}}\right)
\end{equation}
Therefore,
\begin{equation}
    \mathbb{P}\left[\max _{x \in D_{t}}\left|\bar{y}_{t}(x)-f(x)\right|>\frac{\delta}{2}\right] \leq \sum_{i=1}^{N_{t}} \mathbb{P}\left[\left|\bar{y}_{t}\left(x_{i}\right)-f\left(x_{i}\right)\right|>\frac{\delta}{2}\right] \leq 2 N_{t} \exp \left(-\frac{\delta^{2} s_{N_{t}}}{8 \sigma_{0}^{2}}\right)
\end{equation}
Hence,
\begin{equation}
    (\romannumeral1)=\mathbb{P}\left[\left|\bar{y}_{t}\left(\hat{x}_{t}\right)-f\left(\hat{x}_{t}\right)\right|>\frac{\delta}{2}\right]<\mathbb{P}\left[\max _{x \in D_{t}}\left|\bar{y}_{t}(x)-f(x)\right|>\frac{\delta}{2}\right] \leq 2 N_{t} \exp \left(-\frac{\delta^{2} s_{N_{t}}}{8 \sigma_{0}^{2}}\right)
\label{sim_bonud}
\end{equation}

We next bound the term $(\romannumeral2)$. Let $A:=\left\{\left|\bar{y}_{t}\left(\hat{x}_{t}\right)-f\left(\hat{x}_{t}\right)\right|<\frac{\delta}{5}\right\}$, $B:=\left\{\left|\bar{y}_{t}\left(x_{0}^{t}\right)-f\left(x_{0}^{t}\right)\right|<\frac{\delta}{5}\right\}$, and $C:=A\cap B$. Then,
\begin{equation}
\begin{aligned}
    (\romannumeral2)&=\mathbb{P}\left[\left|f\left(\hat{x}_{t}\right)-f\left(x_{0}^{t}\right)\right|>\frac{\delta}{2}\right]\\
    &=\underbrace{\mathbb{P}\left[\left\{\left|f\left(\hat{x}_{t}\right)-f\left(x_{0}^{t}\right)\right|>\frac{\delta}{2}\right\} \bigcap C\right]}_{(\romannumeral3)}+\underbrace{\mathbb{P}\left[\left\{\left|f\left(\hat{x}_{t}\right)-f\left(x_{0}^{t}\right)\right|>\frac{\delta}{2}\right\} \bigcap C^{\complement}\right]}_{(\romannumeral4)}
\end{aligned}
\end{equation}
We first show that $(\romannumeral3)$ is zero by contradiction. When both $A,B$ hold, and $\left|f\left(\hat{x}_{t}\right)-f\left(x_{0}^{t}\right)\right|>\frac{\delta}{2}$, we can conclude that $\bar{y}_{t}\left(\hat{x}_{t}\right)>\bar{y}_{t}\left(x_{0}^{t}\right)$. This contradicts with the fact that $\bar{y}_t(\hat{x}_t)$ is the current best. To bound $(\romannumeral4)$, using a similar inequality of (\ref{sim_bonud}), we have

\begin{equation}
   (\romannumeral4)<\mathbb{P}(C^{\complement})=1-\mathbb{P}(A\cap B)<2-\mathbb{P}[A]-\mathbb{P}[B]<4 N_{t} \exp \left(-\frac{\delta^{2} s_{N_{t}}}{50 \sigma_{0}^{2}}\right)
\end{equation}

Hence, 
\begin{equation}
    (\romannumeral2) = 0+(\romannumeral4) <4 N_{t} \exp \left(-\frac{\delta^{2} s_{N_{t}}}{50 \sigma_{0}^{2}}\right)
\end{equation}

\begin{equation}
\mathbb{P}\left[\left|\bar{y}_{t}^{*}-f\left(x_{0}^{t}\right)\right|>\delta\right] < (\romannumeral1)+(\romannumeral2)<6 N_{t} \exp \left(-\frac{\delta^{2} s_{N_{t}}}{50 \sigma_{0}^{2}}\right)
\end{equation}
By Assumption 1, 
\begin{equation}
\sum_{t=1}^\infty\mathbb{P}\left[\left|\bar{y}_{t}^{*}-f\left(x_{0}^{t}\right)\right|>\delta\right] < 6 \sum_{t=1}^\infty N_{t} \exp \left(-\frac{\delta^{2} s_{N_{t}}}{50 \sigma_{0}^{2}}\right)<\infty
\end{equation}

\subsection{$f(x_0^t)-f(x^*) \to 0$ w.p.1 as $t \to \infty$}
If we assume the mean response $f$ is continuous, then for any $\delta>0$, we can select a region $S$ near $x^*$, such that for all $x\in S$, $|f(x)-f(x^*)|\leq\delta$. From the density of the design points, for any $\epsilon>0$, there is a large $K$ such that there exists at least one design point selected in $S$ before $K$-th iteration. It follows that 
\begin{equation}
    \mathbb{P}\left[\left|f\left(x_{0}^{t}\right)-f\left(x^{*}\right)\right|>\delta \quad i . o .\right]=0
\end{equation}
Hence, $f(x_0^t)-f(x^*)\to 0$ w.p.1 as $t \to \infty$.

\subsection{$\bar{y}_t^* - f(x^*) \to 0$ w.p.1}
Since both $\bar{y}_t^* - f(x_0^t) \to 0$ and $ f(x_0^t) - f(x^*) \to 0$ w.p.1 as $t \to \infty$, from the property of convergence with probability 1, we directly have 
$\bar{y}_t^* - f(x^*) \to 0$ w.p.1, which completes our proof.

\section{Test Functions}
\begin{enumerate}
    \item Branin: $f(x_1,x_2)=a\left(x_{2}-b x_{1}^{2}+c x_{1}-r\right)^{2}+s(1-t) \cos \left(x_{1}\right)+s$, with $a=1,b=\frac{5.1}{4\pi^2},c=\frac{5}{\pi},r=6,s=10,t=\frac{1}{8\pi}$

    \item Camel: $f(x_1,x_2)=\left(4-2.1 x_{1}^{2}+\frac{x_{1}^{4}}{3}\right) x_{1}^{2}+x_{1} x_{2}+\left(-4+4 x_{2}^{2}\right) x_{2}^{2}$

    \item  Eggholder: $f(x_1,x_2)=-\left(x_{2}+47\right) \sin \left(\sqrt{\left|x_{2}+\frac{x_{1}}{2}+47\right|}\right)-x_{1} \sin \left(\sqrt{\left|x_{1}-\left(x_{2}+47\right)\right|}\right)$

\end{enumerate}

\end{appendix}

\end{document}